\documentclass[12pt,man,floatsintext]{apa7}

\usepackage{url}
\usepackage{amssymb}
\usepackage{amsmath}
\usepackage{amsthm}
\usepackage{graphicx}
\usepackage{caption}
\usepackage{subcaption}
\usepackage{comment}
\usepackage{algorithmic}
\usepackage{booktabs}
\usepackage{hyperref}

\usepackage{bbm}
\usepackage[style=apa]{biblatex}
\usepackage{float} 
\usepackage{lineno}
\usepackage{verbatim}

\title{``Just in Time'' World Modeling Supports Human Planning and Reasoning}
\shorttitle{Just in Time World Modeling}

\authorsnames[1,1,{2},1,1]{Tony Chen,Sam Cheyette, Kelsey R Allen, Joshua B Tenenbaum, Kevin A Smith}
\authorsaffiliations{{MIT Department of Brain and Cognitive Sciences},{UBC Departments of Computer Science and Psychology}}

\abstract{Probabilistic mental simulation is thought to play a key role in human reasoning, planning, and prediction, yet the demands of simulation in complex environments exceed realistic human capacity limits. A theory with growing evidence is that people simulate using \emph{simplified representations} of the environment that abstract away from irrelevant details, but it is unclear how people determine these simplifications efficiently.  Here, we present a ``Just-in-Time'' framework for simulation-based reasoning that demonstrates how such representations can be constructed online with minimal added computation. The model uses a tight interleaving of simulation, visual search, and representation modification, with the current simulation guiding where to look and visual search flagging objects that should be encoded for subsequent simulation. Despite only ever encoding a small subset of objects, the model makes high-utility predictions. We find strong empirical support for this account over alternative models in a grid-world planning task and a physical reasoning task across a range of behavioral measures. Together, these results offer a concrete algorithmic account of how people construct reduced representations to support efficient mental simulation.

}

\newcommand{\model}{Just-in-Time}
\newcommand{\shortmodel}{JIT}

\begin{document}


\maketitle
\section{Introduction}

Human planning and reasoning rely on the ability to mentally simulate states of the world, such as imagining a free path through a cluttered kitchen, or evaluating the result of a billiards shot.
However, real-world scenes contain a staggering amount of detail, and simulations of complex environments will necessarily contain far more detail than can be plausibly held in working memory \parencite{garner1953informational,miller1956magical, vul2009explaining, luck2013visual, alvarez2004capacity, vogel2001storage}.
How, then, are people able to fluently plan, predict, and pursue their goals? 

Recent work has focused on the idea that people construct \textit{reduced representations} or \textit{construals} of the environment that abstract away irrelevant details and contain only relevant features. Based on findings that people are selective in what they remember about environments \parencite{shin2023learning,kinney2024building}, there have been attempts to formalize normative principles that describe what the mind chooses to represent. 
These theories often center around the idea that human cognition has limited capacity and must therefore make a rational trade off between the utility that representing an object has for improving our plans or decisions, and the cognitive costs of that representation \parencite{gershman2015computational,lieder_resource-rational_2020,ho2022people,arumugam2024bayesian,icard2015resource}.

However, while there has been extensive research on \emph{what} people should represent, there has been little focus on \emph{how} these construals are formed. A number of popular accounts of simulation-based reasoning implicitly assume that representation formation is 
a process that occurs \textit{prior to} simulation or planning, and the utility of a construal is defined as how closely a simulation under that construal matches the simulation that would have been produced considering all objects.
Yet this formulation leads to a paradoxical conclusion: exactly evaluating the utility of a construal requires considering a simulation using the full environment, making the search for an optimal construal more computationally taxing than simply including all objects in the first place \parencite{russell1991principles}.
While there have been proposals that avoid this paradox by, for example, learning to approximate utility based on prior experience \parencite{lieder2017strategy} or finding the best resource-limited plan within a time budget \parencite{gershman2015computational}, these strategies do not easily generalize to novel tasks, or may waste computation on simulating construals that are subsequently discarded.



Here, we propose a formal model of simulation-based reasoning in which representations of the environment are incrementally constructed and modified. Our \model~(\shortmodel) model uses a local visual scan to quickly flag high expected-utility objects relevant for future steps of simulation. Details about the environment are added to a construal only when they are relevant to immediate next steps of the current simulation.
In contrast to proposals that cast construals as the result of an explicit optimization \parencite[e.g.][]{ho2022people,icard2015resource} --- which require running many simulations up front to evaluate candidate construals --- in our account, a construal is the result of iteratively adding relevant objects to memory throughout a single simulation. The \shortmodel~model thus avoids the conceptual issues that arise from assuming that construals are built or evaluated prior to running simulations, while simultaneously finding construals that support accurate predictions or plans with minimal representational demands. In fact, perhaps surprisingly, the \shortmodel~model produces representations that on average contain \emph{fewer} objects than models that assume precomputation, without a loss of performance in either planning or physical reasoning, and without requiring many simulations to evaluate the utility.
We test the \shortmodel~model as an account of human simulation-based reasoning in two domains: a grid-world planning task introduced in \textcite{ho2022people}, where participants navigate a grid-like maze, and a physical prediction task similar to \textcite{gerstenberg2018happened}, where participants predict the trajectory of a ball as it falls through an array of obstacles.
We focus specifically on the domains of planning and physical reasoning for three reasons. First, these are ecologically important domains in which people must function well in cluttered environments.
Second, although planning and physical reasoning are often treated as distinct cognitive domains, prior research indicates that both capacities rely on mental simulation to select actions based on their expected outcomes \parencite{battaglia_simulation_2013,gerstenberg2021counterfactual,van2023expertise}, suggesting commonalities in mental computation that our model can exploit. Finally, these domains nonetheless differ in a number of important ways --- including both their objective dynamics and the brain areas supporting them \parencite{fischer2016functional,pramod2022invariant,koechlin2000dissociating,tanji2008role} --- which allows us to test the generality of our framework. Altogether, our results provide evidence that people construct simplified representations of scenes by interleaving representation formation with simulation, producing effective and efficient construals to support planning and reasoning.
\section{The \model~model}



We propose the \model~(\shortmodel) model as a model of memory-efficient and time-efficient mental simulation that selectively represents a subset of the scene.
Inspired by ``just-in-time'' principles---implemented across a variety of fields including computer science \parencite{aycock2003brief} and logistics \parencite{pisch2020managing}---that defer tasks until the results are needed, the \shortmodel~model defers representing objects until just before that representation is relevant for the current simulation.
The framework we describe is general, requiring only three interacting components.
First is a \textit{representational sketchpad} that contains object-centric information to support simulation.
Next is a \textit{simulator} that, given a current state $s$ and optionally an action $a$, probabilistically forecasts possible future states: $P(s'|s,a)$.
Finally we assume a \textit{perceptual lookahead} module that uses the current state of the simulation to guide a local visual search to scan the scene for potential upcoming interactions and objects to include in the representation.
The outcome of this iterative process is a \textit{construal} of the scene: the final contents of the representational sketchpad including the set of objects whose effects are modeled in simulation: $C \subseteq O$.
We assume that memory is fallible, so that the longer a previously relevant object has not been used, the higher the probability it is forgotten. The construal arises from an iterative process of simulating, looking ahead, and encoding, each step of which we detail next.



\begin{figure}[t!]
    \centering
    \includegraphics[width=0.99\linewidth, trim=0cm 0cm 0cm 0cm, clip]{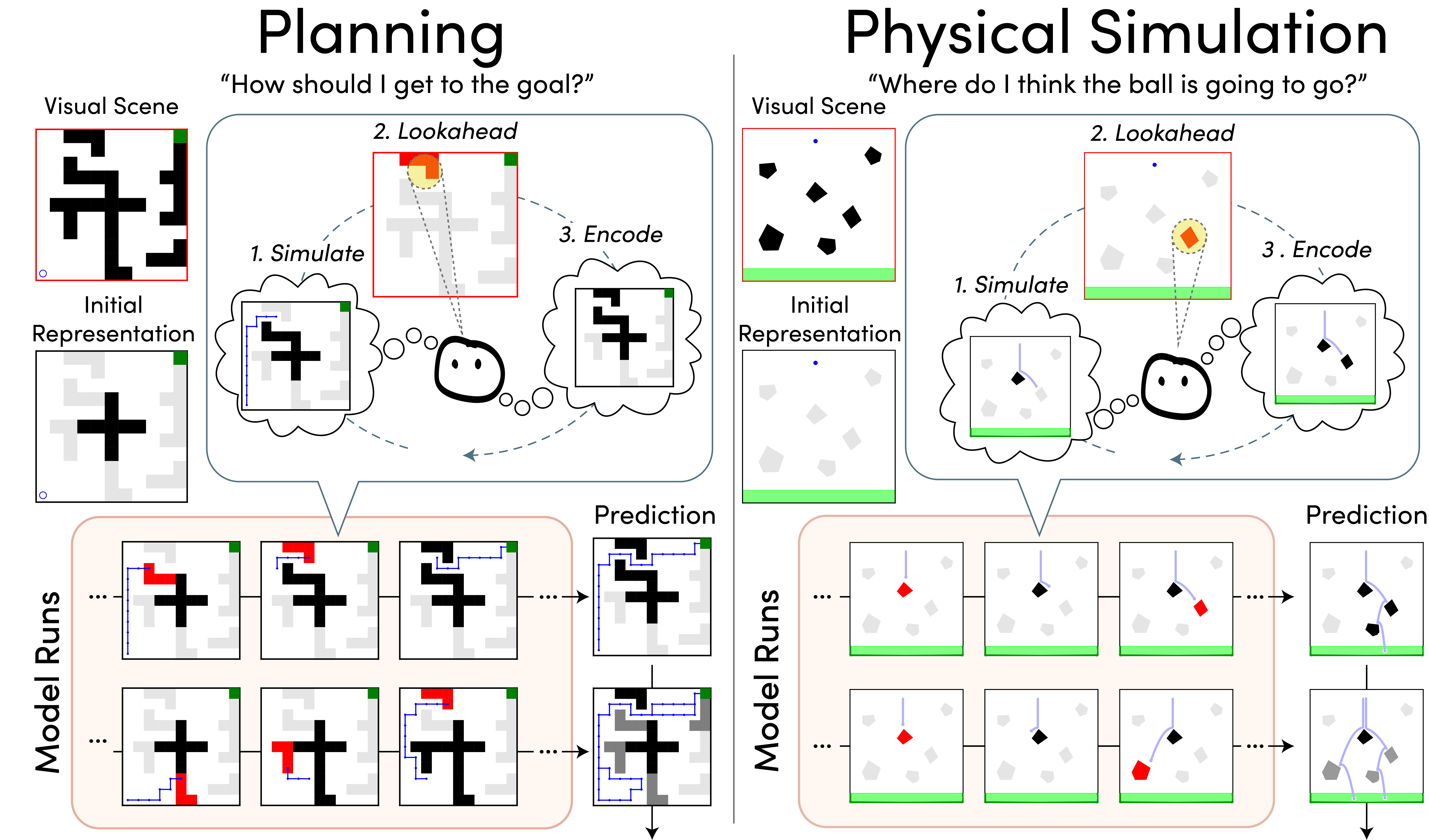}
    \caption{The  \model~model for  constructing representations for planning and reasoning, and the two test domains. \textbf{Left:} a grid world task where the goal is to navigate an agent from start (blue circle) to goal (green square). \textbf{Right:} a physical prediction task, where the goal is to predict where a ball will land after falling through an array of obstacles. \shortmodel~incrementally updates a representation in between steps of simulation and perceptual lookahead.
    At each step of a model run, the state of the simulation is first incremented (simulate). Then, this state is used to guide a visual search of the scene to find potential collisions (lookahead), and any objects flagged by this search are then used included in the representation (encode), to drive the next steps of simulation.
    In the planning domain, simulation is implemented as a stochastic variant of the $A^\star$ search algorithm \parencite{zhi2020online}, and in the physics domain, simulation is implemented as a probabilistic physics simulation engine \parencite{battaglia_simulation_2013}
    }
    \label{fig:intro}
\end{figure}


\noindent \textbf{Initializing the representation.}
The tasks that we consider require tracking or moving a single object through a mostly static environment. The representation is therefore initialized sparsely such that it only includes the tracked object, goal areas, and features common to all scenarios. 


\noindent \textbf{Simulate.}
At each step, the state is incremented by sampling from the simulator's probability distribution: $s' \sim p(s'|s,a)$. The collection of states sampled so far produces a \textit{trajectory}, which we label as $\tau = (s_1, s_2, \ldots s_n).$
\shortmodel~produces predictions by sampling many such trajectories and aggregating their results.

\noindent \textbf{Lookahead.}
After taking a simulation step, the model scans the visual scene for potential interactions that may be imminent, to identify objects that are not yet represented, but need to be.
People's eyes naturally follow the simulated trajectory of objects \parencite{gerstenberg2017eye,callaway2024revealing,huber2004ball}, and by this tracing process we assume they will look at objects that are along this simulated path.
This procedure draws inspiration from accounts of visual attention that describe the environment as a kind of external memory \parencite{o1992solving}:
even if the simulated trajectory unknowingly violates physical constraints like interpenetration, the model can still check the trajectory against the visual scene to find an error and identify the offending object.


\noindent \textbf{Encode.}
After lookahead, the representation is first updated with the new objects flagged by the lookahead module: $C \leftarrow C \bigcup \{o: o \in \ell(s)\}$. Then, the representation is culled, probabilistically forgetting objects that have not been relevant for some time.
Intuitively, objects that are encountered early in simulation often do not need to be revisited; and so continuing to encode an object that was previously represented but is not currently needed is wasteful.
We therefore assume that objects are stochastically dropped from the representation with probability according to a power law \parencite{anderson1989human}, controlled by a decay parameter $\gamma$ ($p(\text{forget object } o) \propto t^{-\gamma}$,
where $t$ is the number of steps since object $o$ has been encoded).


\subsection{Domains for testing the \shortmodel~model}

We test the \shortmodel~model in two domains to demonstrate its generality: planning and physical reasoning. The tasks we use share two key features. First, relevant interactions in both cases are always \emph{spatially local}---i.e., objects require proximity to affect each other. This allows for efficient lookahead, which only needs to consider a limited set of prospective objects that are close to the simulation trajectory. Second, scenes are always \textit{perceptually available} throughout the task, allowing objects to be accessed on demand by querying the external visual scene.



\subsubsection{Planning}

We use the same task as \textcite{ho2022people} for the planning domain (see Fig.~\ref{fig:intro}, \emph{Left}). An agent is placed in a grid world and must move their avatar to a green goal. However, there are walls blocking the movement of the avatar -- there is always a cross in the middle of the world, and a number of other L-shaped walls that move between scenarios.

Because the avatar and goal are task-relevant, and because the cross is always in the same location, representations are initialized with these three objects; however, all other walls must be explicitly added. Based on evidence that people plan stochastically instead of deterministically maximizing utility when navigating \parencite{zhu2015people}, we implement simulation using a stochastic variant of an $A^\star$ planner \parencite{zhi2020online}.
The $A^\star$ planner uses a Manhattan distance heuristic and a straight line tiebreak bias, such that when reconstructing a plan from the sequence of visited states, the planner will favor plans that travel in a straight line.

\subsubsection{Physical reasoning}

For the physical reasoning task, we consider a variant of the task introduced in \textcite{gerstenberg2018happened}: a blue ball is dropped through a number of bumper objects and the goal is to predict where the ball will land on the green floor (see Fig.~\ref{fig:intro}, \emph{Right}). The model's representation of the scene initially contains only the ball and the floor; the bumpers must be added via \shortmodel. We assume that simulation is probabilistic, including uncertainty about the initial position of the ball as well as noise in how collisions resolve \parencite{allen_rapid_2020,smith2013sources,wang2024probabilistic}.

\subsection{An alternate model: Value Guided Construal}

We compare our \shortmodel~model to the \emph{Value Guided Construal} (VGC) model proposed by \textcite{ho2022people}. VGC proposes that people should choose the construal that jointly optimizes the utility that can be obtained by planning using this construal, while minimizing the number of objects contained in that construal.
%
This objective formalizes the parsimony of a representation: excluding an important object leads to infeasible plans or impossible physics, while representing irrelevant objects increases the complexity of representation without any gain in value.

While both models approach the construction of a construal from resource-rational principles, a key difference is in how \shortmodel~and VGC treat the utility of a construal. \shortmodel~predicts that objects should be represented with a strength proportional to ``need probability'' \parencite{anderson1989human} -- how likely that each object is to be used in simulation -- while VGC directly optimizes the representational strength of each object to maximize the utility of a planner or simulator using that representation.
Thus there are two key scenarios in which \shortmodel~and VGC will differ:

\begin{itemize}
    \item There are two equally likely initial plans, but one is blocked by later objects in the environment. In this case VGC will always represent the blocking object as it affects planning overall, but \shortmodel~will only represent the blocking object if a given plan is initialized in that direction (see Extended Data Figure \ref{fig:extended-theory-example}).
    \item A set of objects changes the trajectory of simulation, but in such a way that it does not change the ultimate outcome. In this case, \shortmodel~will represent those objects as they become relevant to the simulation, whereas VGC will ignore the objects since including them will not change the utility of the simulations (see Fig.~\ref{fig:physics-results}B)
\end{itemize}

\textcite{ho2022people} propose four variants of the VGC model, so to compare against \shortmodel, we use their best fitting variant. This variant assumes that decision-makers modify their construal throughout the process of making their plans. Thus the best construal is the weighted average of the optimal construals at each point in the plan. In effect, this predicts that objects that are close to the start of a plan will have lower memory strengths as they are only relevant to the plan for short periods of time, whereas objects nearer to the goal will be better represented, as those objects are relevant throughout the plan.

For further details on the implementation of VGC, see \textcite{ho2022people} and the Supplementary Materials Section~\ref{sec:supp-vgc-phys}.

\section{\shortmodel~Explains Human Planning}

\subsection{Representations for planning}

\begin{figure}[htp]
    \centering
    \includegraphics[width=0.7\linewidth]{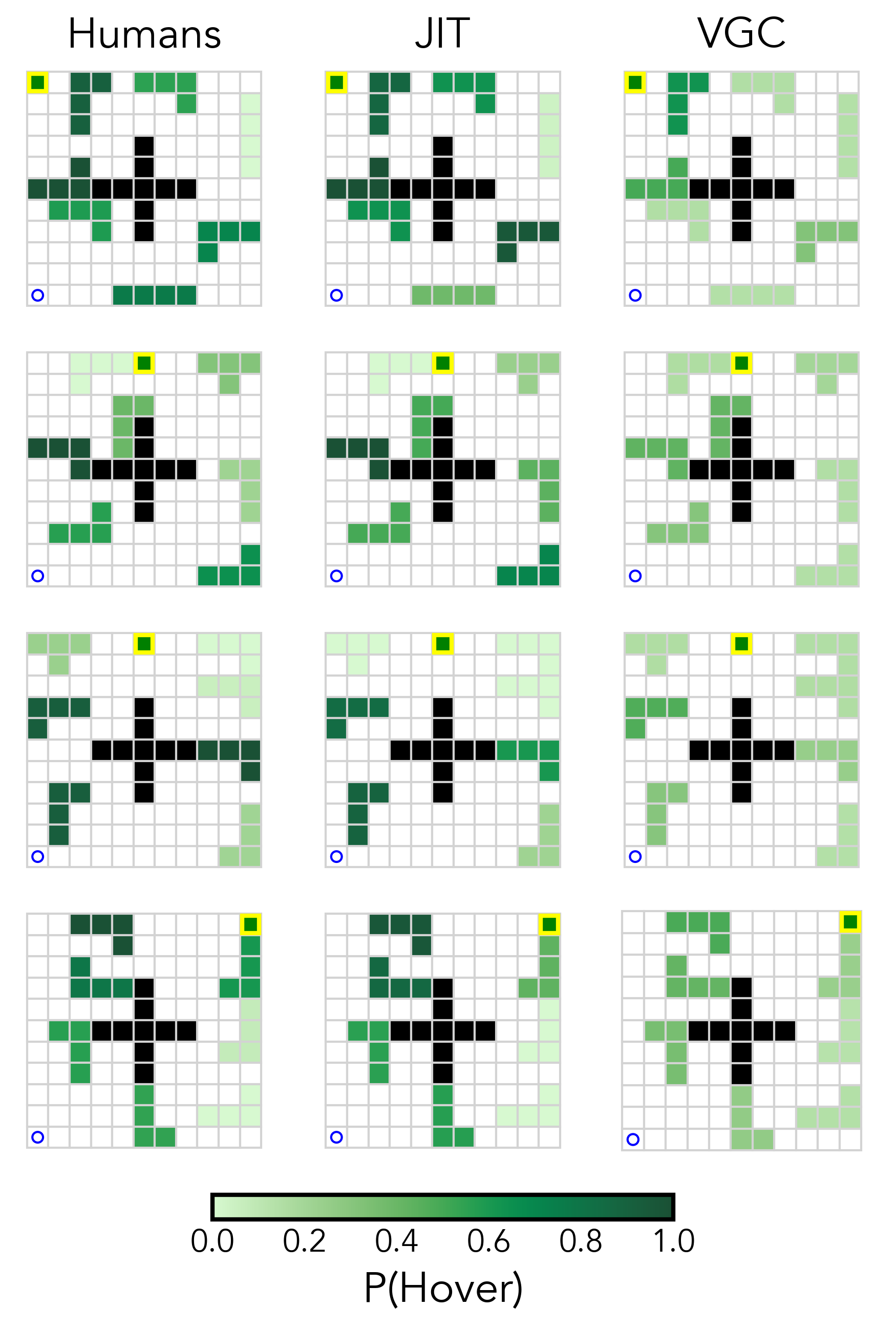}
    \caption{
    Comparisons between human results, \model~model, and the Value-Guided Construal model in the process-tracing experiments (1D and 1E) of \cite{ho2022people}.
    Participants were shown a blank maze in which only the center object was initially visible. They were instructed that moving the mouse over a masked object would reveal it.
    Their task was to plan a path from the start (blue circle) to the goal (highlighted green square), revealing as many objects as needed along the way.
    Heatmaps show the probability of an object being revealed by people, and the representational weight \shortmodel~and VGC assign to each object.
    }
    \label{fig:navigation-heatmaps}
\end{figure}

The primary experimental evidence for value-guided construals comes from measures of memory and attention in the grid-world navigation task of \textcite{ho2022people}. We re-analyze the behavioral data from all 9 experiments in that paper (Experiments 1A-I). In each experiment, the task consisted of two phases, a ``planning phase'' and an ``execution phase.''
In the ``planning phase,'' participants were first shown an image of a grid-world --- like the one shown in Figure~\ref{fig:intro} (left) --- and asked to plan a path from start to goal.
In the ``execution phase,'' obstacles were masked out and the participant was asked to execute their plan by moving the agent to the goal using a keyboard.

We highlight results from three experiments in their paper (see the Extended Data Figure~\ref{fig:extended-navigation-scatterplots} for comparisons across all experiments): one that probed human memory (1C) in a way that is directly comparable to our physical prediction experiments, and two that probe attention (1D and 1E), as these experiments most directly test the dynamics of representation formation that we expect \shortmodel~to capture. 
In Experiment 1C, participants' memory for the location of objects was probed immediately after generating a plan in the execution phase.
Participants were shown the original world except for one object replaced with two colored objects placed in different positions: one in the original position of the object and another shifted into a nearly position. They were then asked to identify which object was in the original, correct position and how confident they were on an eight-point scale.  In Experiments 1D and 1E, which aimed to capture the dynamics of human attention, the objects were masked out during the planning stage and could only be revealed by hovering over them with a mouse. The dependent measure in these experiments was the proportion of participants who hovered their mouse over that object for any amount of time in the planning phase.


Results showing the fit of both the \shortmodel~model and the VGC model to human data from 1C-E are shown in Figures \ref{fig:navigation-heatmaps} and \ref{fig:navigation-results}A.
In experiment 1C, the \shortmodel~model has a higher correlation, lower RMSE, and higher log-likelihood than VGC (\shortmodel: $r=0.95$, $\text{RMSE}=0.08$, $LL\text{: }-1,763$; VGC: $r=0.93,\text{ RMSE}=0.10$, $LL\text{: }-1,809$) while having one fewer free parameter. However, the predictions from the \shortmodel~model and the VGC model were highly correlated ($r=0.95$). We also compare the probability of an object being included in a construal to the probability an object was hovered during the planning stage in Experiments 1D and 1E.
Because there is no influence of memory on the probability of an object being hovered, we fix the \shortmodel~memory decay parameter $\gamma$ to 0. 
Again, we find that our model explains human information seeking better than the value-guided construal solution in both Experiment 1D (\shortmodel: $r=0.88$,  RMSE $=0.18$, $LL = -8,093$; VGC: $r=0.65,\text{ RMSE}=0.40$, $LL=-11,864$) and Experiment 1E (\shortmodel: $r=0.95,\text{ RMSE}=0.14,~LL=-5,892$; VGC: $r=0.74$,\text{ RMSE}=0.29, $LL=-7,607$).






\subsection{
The Efficiency of \shortmodel~Representations
}
\label{sec:efficiency}

\begin{figure}[htp]
    \centering
    \includegraphics[width=0.77\linewidth]{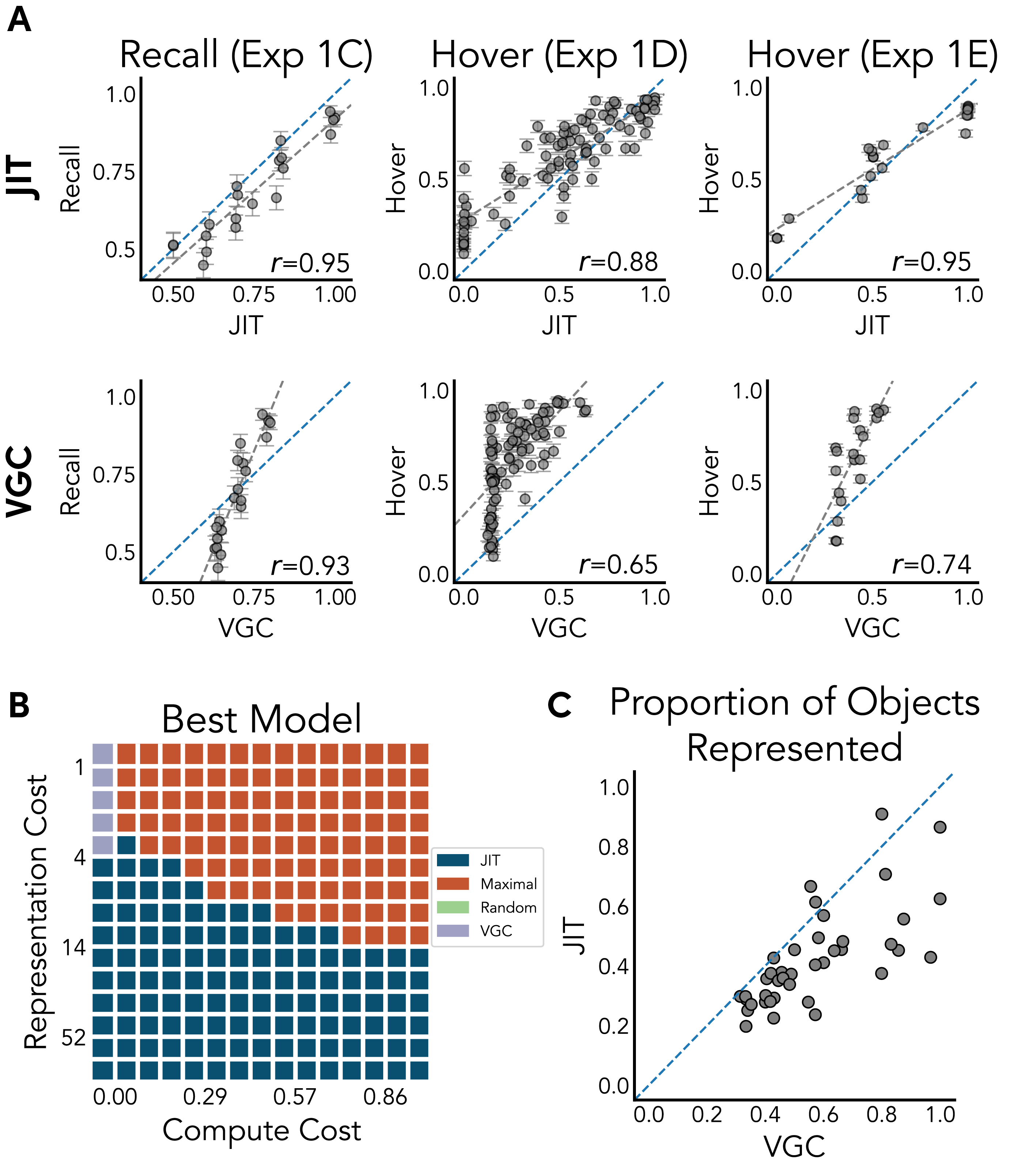}
    \caption{ 
    \textbf{(A)} Predictions from our \shortmodel~model (top row) compared to value-guided construal (bottom row) against human responses for three experiments in \textcite{ho2022people}. The blue line denotes the identity line, and the gray line denotes the line of best fit.
    \textbf{(B)} 
    Quantitative model comparisons of compute and memory resource use for a set of procedurally generated grid worlds.
    We calculate a resource efficiency measure for four models (\shortmodel~compared to planning with the maximal representation, planning with a random representation, and planning with VGC) under a variety of cost assumptions. The net utility is inversely proportional to the sum of costs of representation (y-axis), computation (x-axis), and the length of the resulting plan. Heatmap colors visualize the most efficient model for each parameter regime. 
    \textbf{(C)} Average proportion of objects represented by \shortmodel~and VGC in the same procedurally generated worlds.
    }
    \label{fig:navigation-results}
\end{figure}

Next, we conduct a series of simulation analyses to analyze the efficiency of representations that \shortmodel~finds in the planning task. We procedurally generate a set of 40 random grid worlds with a random number of objects, and analyze both the complexity of representations found by a set of models, and the computational overhead of constructing these representations.
To do so, we compute an \textit{algorithmic utility} $V$ that, given an algorithm $A$, trades off the utility of planning, the complexity of representation, and the computational cost of planning, and then score \shortmodel~along with a series of baselines on this measure:


$$
V(\text{A}) = \mathbb{E}[U(\text{A})] - \alpha \cdot C_{\text{computation}}(\text{A}) - \beta \cdot C_{\text{representation}}(\text{A}).
$$

Above, the utility of planning with an algorithm A is the negative length of the found plan, the cost of computation is the number of nodes expanded during search, and the cost of representation is the number of objects included in the construal.
We compare \shortmodel~against three alternative models: a \textit{maximal} baseline that always plans with all objects included, a \textit{random} baseline that plans with a randomly selected construal, and a modified \textit{VGC} model adapted to use heuristic search instead of policy iteration, so that the measure of computational cost can be consistent across all models.
For each $\alpha$ and $\beta$ within a wide range, we find the model with the highest utility, and visualize the results in Fig.~\ref{fig:navigation-results}B. 
We find \shortmodel~dominates in most regimes with two exceptions: VGC wins when there are no computational costs at all and only a low cost of representation ($\alpha$ is $0$ with low $\beta$), and the maximal model performs best when computation is expensive and representation is cheap (high $\alpha$ and low $\beta$).

A key feature of \shortmodel~is that it computes a representation that supports the currently simulated trajectory, and not the representation necessary for planning the optimal trajectory.
In many cases, an object might be deemed relevant by VGC because its existence changes an otherwise optimal path into a suboptimal one, but according to \shortmodel, may be bypassed entirely during planning if the initial plan happened to take a different path.
This leads our model to represent \textit{fewer} objects than VGC on average, as shown in Fig.~\ref{fig:navigation-results}C (see also Extended Data Figure~\ref{fig:extended-theory-example}).
So while VGC produces optimal construals \emph{under the assumption that all planning must be performed with that representation}, \shortmodel~only considers objects relevant to the singular \emph{current} plan and will therefore often ignore objects in the optimal construal for VGC.

\begin{figure}[tp!]
    \centering
    \includegraphics[width=0.9\linewidth,trim={0cm 0cm 0cm 0cm},clip]{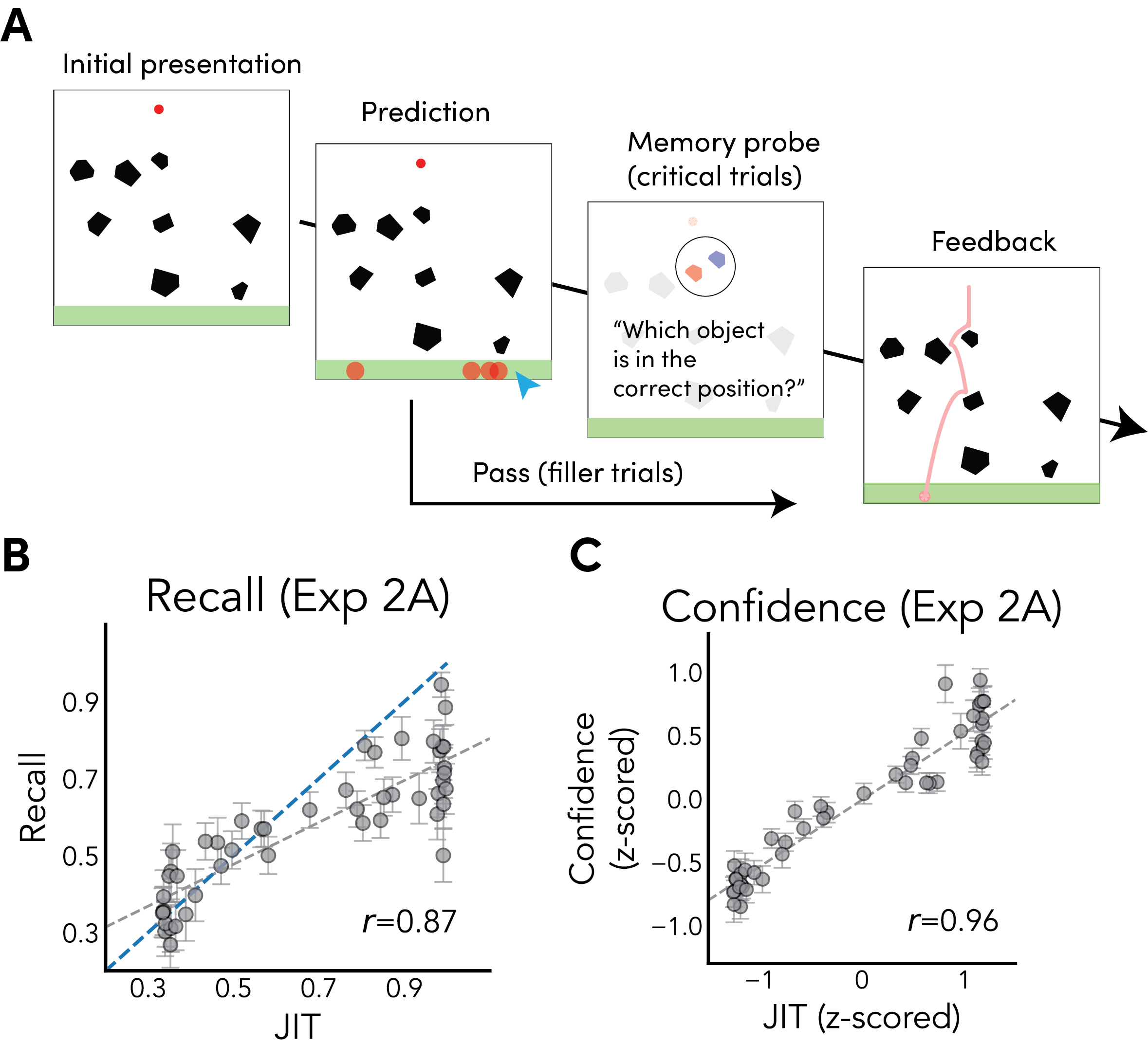}
    \caption{\textbf{(A)} Design of Experiment 2A. Participants saw a red ball above an array of obstacles and made predictions about where it would land. On a subset of trials, participants were then subsequently probed for their memory of a randomly chosen object's position.
    \textbf{(B)} Scatterplot of model predictions against human memory for objects. The blue dashed line denotes the identity line, and gray dashed lines denote the line of best fit. Error bars denote standard error. \textbf{(C)} Scatterplot of model predictions against confidence in recall (both z-scored).}
    \label{fig:physics-experiment}
\end{figure}



\section{\shortmodel~Explains Representations for Physical Prediction}

While we have demonstrated that \shortmodel~can explain the representations that people form during planning, we have proposed this as a general cognitive framework people use for constructing representations for simulation-based reasoning.
We therefore next show how \shortmodel~predicts human representations in the domain of physical prediction as a demonstration of the generality of this framework.

\begin{figure}[htbp]
    \centering
    \includegraphics[width=0.75\linewidth]{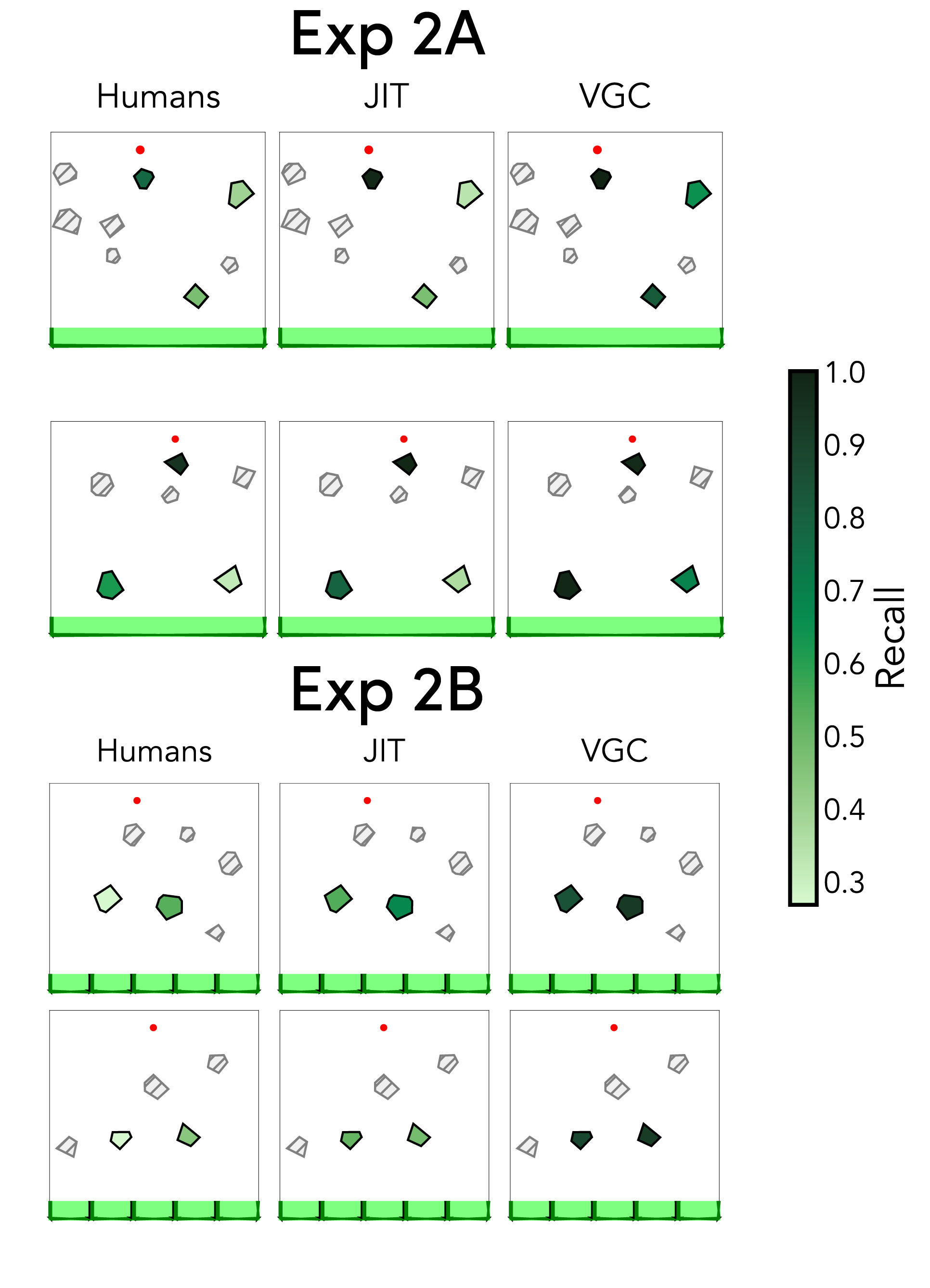}
    \caption{
    Selected worlds from Experiment 2A and 2B, where participants predicted the path that a red ball will take when let go, and were subsequently probed for memory of selected objects in the scene. We fit model parameters to maximize correlation with recall from experiment 2A, and transfer those parameters as-is to experiment 2B. Heatmaps show human recall for probed objects and predicted representations from both \shortmodel~and VGC; we did not probe memory for the gray hatched objects.}
    \label{fig:physics-heatmaps}
\end{figure}

\subsection{Representations for physical prediction}


We adapt the Plinko environments originally proposed in \textcite{gerstenberg2018happened} as our primary experimental domain.
An example of the task is shown in Fig.~\ref{fig:physics-experiment}A: given a static image of a scene with a ball suspended in mid-air, the goal is to predict where the red ball will land once gravity is turned on.
We use memory probes to externalize the representations of objects that people are forming in these tasks; assuming that objects that are more deeply or frequently encoded during simulation are remembered better \parencite{craik1975depth}.

In Experiment 2A, participants were presented with a Plinko world, and made predictions by clicking on the spots where they think the ball is likely to land. On critical trials (comprising of a third of all trials), participants were probed for their memory of randomly selected objects in the world: we showed participants the original world and a distractor world with the probed object shifted by a random translation, and asked them to identify which world was the correct one that they had just made predictions for.  Judgments used a sliding scale sliding scale with extreme ends indicating surety in one of the two objects being in the original world, and the middle labeled as uncertain about the original location. We transformed these responses into `recall' measures indicating the proportion of participants having any evidence for the correct object, and `confidence' measures indicating the average position of the slider as a proportion between the middle and the end.


\subsubsection{Comparing the \shortmodel~model against people}

We next consider how well our model can predict the exhibited memory patterns, by correlating the computed representation traces of our model against the average recall for a given object in a given Plinko world.
Our model has four total free parameters: three noise parameters that govern the behavior of the probabilistic simulator, and the decay parameter that governs how quickly object traces fade over time. We fit the noise parameters to a separate norming experiment (see Supplementary Section \ref{sec:supp-param-fit}) and fit the decay parameter to maximize the correlation between model predictions and human recall.

We find a strong correlation between the representation trace of our model and both recall ($r=0.87,\text{RMSE}=0.18$; Fig.~\ref{fig:physics-experiment}B) and confidence ($r=0.96$; Fig.~\ref{fig:physics-experiment}C).
This effect is not driven by individual variation among participants; 
we find that JIT is a good explanatory factor beyond accounting for individual differences in recall
$(\chi^2(1)=310.33,~~p\approx0)$.
When fitting free parameters of the VGC model to human recall, we find that VGC has a slightly lower correlation to human data, and higher error: (recall: $r=0.82,\text{RMSE}=0.28$, confidence: $r=0.90$).
Similar to the planning memory experiment (1C), JIT and VGC predictions are highly correlated with each other ($r=0.93$), but partial correlations show that JIT explains substantial variation on top of VGC ($r_{\text{JIT|VGC}}=0.49$), while VGC barely improves on \shortmodel: ($r_{\text{VGC|JIT}}=0.08$; see Fig.~\ref{fig:physics-heatmaps}, Extended Fig.~\ref{fig:extended-plinko-modelcomparison}, and Supplementary Section~\ref{sec:supp-vgc-phys} for further detail).
Furthermore, additional control experiments suggest that participants' memory is not driven by low level visual processes such as passive fixation (Supplementary Section \ref{sec:supp-attention}) or scene layout statistics (Supplementary Section \ref{sec:supp-visual}).

\begin{figure}[htbp]
    \centering
    \includegraphics[width=0.99\linewidth]{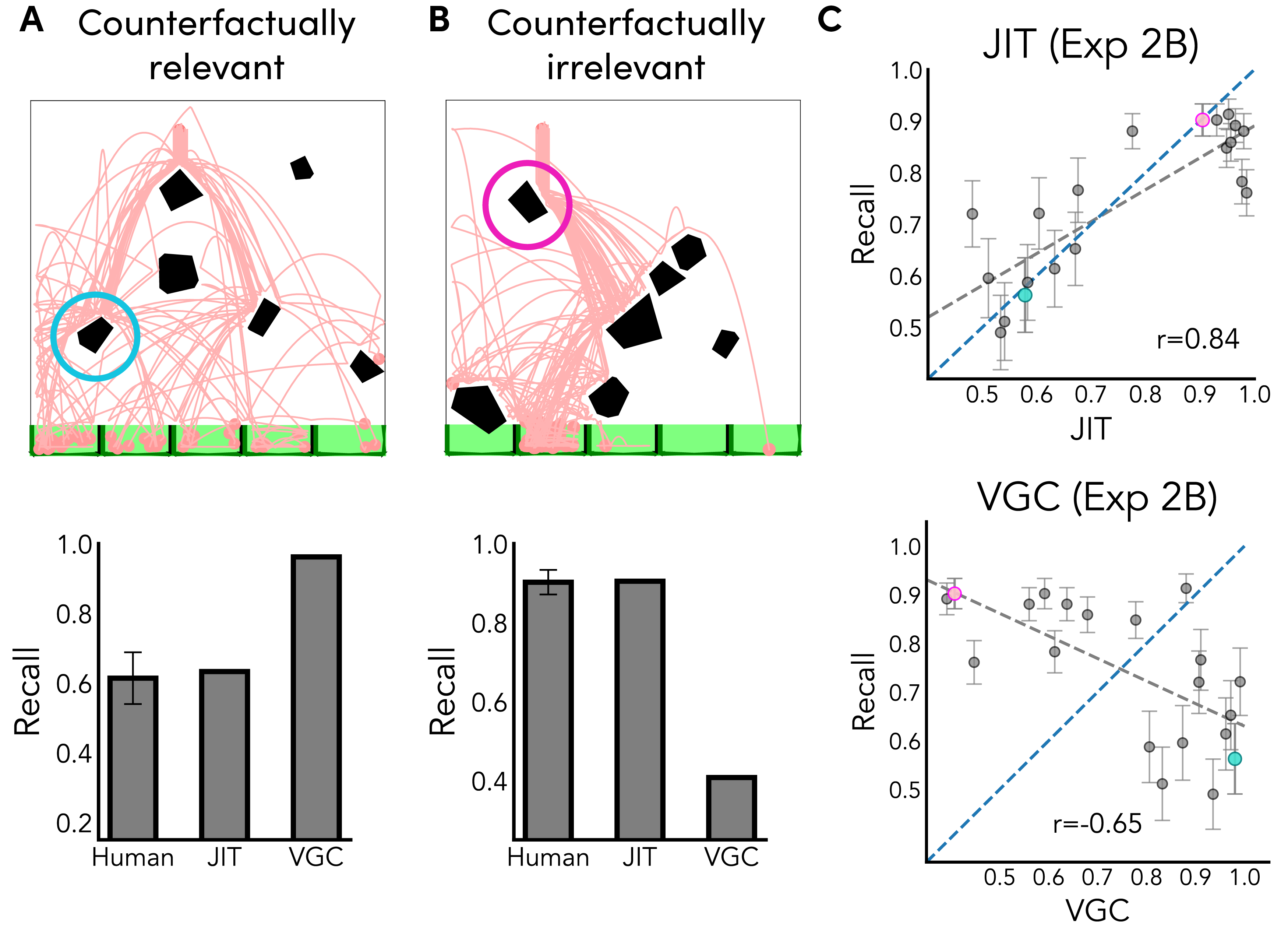}
    \caption{Experiment 2B trials with \textbf{(A)} counterfactually relevant  and \textbf{(B)} counterfactually irrelevant  objects. Light red lines indicate noisy simulations of the ball's path. Bar plots represent recall for humans vs. the \shortmodel~and VGC models for example objects of each type. \textbf{(C)} Correlations of model vs. human recall for \shortmodel~(\emph{top}) and VGC (\emph{bottom}). The blue line represents the identity line, while the grey line is the best fitting line. Error bars indicate standard error. Colored dots indicate data representing the objects shown in A \& B. Like humans, the \shortmodel~model predicts high recall for the counterfactually irrelevant objects and moderate recall for the counterfactually relevant objects, while the VGC model makes the opposite predictions.}
    \label{fig:physics-results}
\end{figure}

\subsubsection{Experimentally distinguishing VGC and \shortmodel}

Experiment 2A showed that the \shortmodel~model was able to explain patterns of human recall for a physical prediction task, but \shortmodel's behavior was highly correlated with that of VGC.
We therefore designed Experiment 2B to include stimuli that would explicitly dissociate the predictions of the two models.

In this experiment, the procedure was the same as Experiment 2A except that in the prediction phase, the ground was segmented into five equally sized buckets, and participants selected which bucket they believed the ball would fall into. These buckets were used to make the outcomes of prediction more granular. We constructed critical worlds containing two types of objects where VGC and \shortmodel~made opposite predictions. 
``Counterfactually Relevant'' objects make a large difference to the final prediction, but are contacted only half of the time under noisy simulation (Fig.~\ref{fig:physics-results}A), and ``Counterfactually Irrelevant'' objects are contacted by the ball but made no difference to the bucket the ball would land in (Fig.~\ref{fig:physics-results}B). VGC predicts that counterfactually irrelevant objects have low value and should not be represented at all, while predicting that counterfactually relevant objects should be represented more strongly. Conversely, \shortmodel~predicts that counterfactually irrelevant objects should be represented to a higher degree than relevant objects, as they show up more frequently in simulation. 

We reuse the exact same model parameters fit to Experiment 2A in order to compare the generality of each of the models. \shortmodel's predictions align well with human memory $(r=0.84,\text{RMSE}=0.11$, see Fig.~\ref{fig:physics-results}C), while VGC predicts the opposite pattern as people $(r=-0.65,\text{RMSE}=0.31)$.
\section{Discussion}

In this work, we have shown how efficient representations of the environment can be constructed without substantial computational overhead by interleaving encoding and simulation. 
We presented the \model~model, a general framework for simulation-based reasoning and planning in visually grounded domains. Just-in-Time simulation effectively balances the trade-off between accuracy, computation, and memory complexity, and also empirically predicts traces of human memory better than competing resource rational models in both a grid-world planning and physical prediction task. These findings support the hypothesis that people incrementally build up a representation of the world \emph{as they simulate}, rather than forming a fixed representation \emph{prior to} simulation. While the intractability and memory burden of simulation in complex environments has been used to argue against the use of mental simulation in physical reasoning \parencite{ludwin-peery_limits_2021,davis2015scope}, our results demonstrate how mental simulation can support physical reasoning and planning without incurring unreasonable memory demands.

The \model~model is able to explain a variety of behavioral measures in two seemingly disparate domains, planning and simulation, with only a single parameter unrelated to the simulator (memory decay). The simplicity of the model, though, is in part enabled by properties of the tasks that make them less complex than most naturalistic scenarios. These simplifications include the fact that the displays in both tasks are static, they involve only a single target object, and they contain at most a handful of relevant objects---each of which is untrue of many real-world scenes \parencite{davis2015scope,battaglia_simulation_2013}.
Incorporating explicit capacity limits and more detailed mechanisms of memory decay may become important in explaining human simulation-based reasoning in cases where there are orders of magnitude more objects involved. The instructions also make clear what the relevant target object is from the outset, and people can fail to notice relevant moving objects if not explicitly cued \parencite{bass2022partial}. A more advanced \model~approach would be needed to account for how people initialize their representations in arbitrary scenes and tasks, relying on pragmatics, visual attention, and prior experience with similar tasks.

Our current investigation has focused on the differences between \model~representation building and pre-computed efficient representations in unfamiliar scenes. However, in many real world situations we do have prior knowledge that can help guide construals prior to any simulation.
Human reasoning in familiar situations may be best explained by a combination of local lookahead and construction of construals through metacognitive deliberation or learned heuristics. For instance, when planning a path through your hometown, you might bring to mind an efficient set of landmarks on the way to your goal based on past experiences traveling to nearby locations, relying on preexisting knowledge that is not immediately available in the current tasks. Future work should study how the local lookahead of \model~simulation and the pre-computed representations of value-guided construals might be efficiently combined, allowing for process models of construal formation in a wider variety of situations.


Nonetheless, our results offer a solution for designing agents that must plan and interact with complex, real-world environments. Modern robotics planning algorithms often face computational challenges when operating in cluttered environments with irrelevant objects \parencite{silver2021planning}.
Similar deferred approaches have been applied to robotics problems such as motion planning \parencite{hauser2015lazy}, but have primarily been analyzed through the lens of reducing computational cost rather than representational complexity.
An incremental approach to constructing representations as an agent plans provides a method for producing simplified construals and therefore reducing the branching factor of planning.

We have presented \model~simulation primarily as a model of cognition in visually-grounded and visually-persistent domains, as these domains allow for cheap and effective implementations of lookahead through visual foveation.
However, the \model~approach is not fundamentally limited to settings involving explicit perceptual input. For instance, local goal-directed search may equally apply to visual imagination, which could explain why people conceptualize only a subset of scene features \parencite{bigelow2023non}. Incremental representation building from local search may also help explain human planning and reasoning in more abstract cognitive domains, such as sub-goaling and subproblem construction in logical reasoning \parencite{olieslagers2024backward,cheyette2025decompose} and foraging-like search patterns in semantic retrieval \parencite{smith2013multiply, hills2012optimal}. \model~representation building may therefore constitute a general strategy by which minds manage complexity and thus become increasingly relevant as we seek to scale models of natural and artificial intelligence to more naturalistic environments.

\section{Methods}

\subsection{Model implementation}

The \shortmodel~model incrementally builds up a representation consisting of relevant objects in the environment during simulation. The output of this model is a \textit{construal}, containing all objects represented in working memory at the end of simulation.
For all analyses, we use the \textit{average} construal as computed using Monte Carlo simulation, which describes the probability an object is present in working memory at the end of simulation.

In the rest of this section we provide full details on how the model is implemented for each of the two domains.
To adapt the model for each individual domain, we must specify the state space $S$; the dynamics of the environment, encoded in a transition function $f$; and how the agent looks for potential collisions, expressed in a lookahead query $\ell(s)$ that takes in the current simulation state and returns any possible colliding objects.

\subsubsection{Planning}

At a high level, our model implementation consists of alternating steps of (1) following a planned path to the goal, and (2) re-planning whenever an obstacle is observed to impede the execution of the plan.
The state of the model $s = (x, y, \pi)$ is given by the current position of the agent, marked by a horizontal and vertical position $x, y$, and the currently active planned path, denoted by $\pi$.
We represent plans as sequences of desired states: $\pi = [(x_0, y_0), \ldots (x_n, y_n)].$
Following Ho et al. \parencite*{ho2022people}, we initialize the construal to contain the center cross obstacle at the beginning of every simulation.

\textbf{Dynamics}. A simulation proceeds by following the computed plan until either the goal is reached, or a possible collision is flagged.
We assume no sources of motor or perceptual error, so that the intended transitions are always reached.
At each step, the simulated agent transitions to the next state specified by the plan, so that $f(x, y, \pi) = (\texttt{next}(\pi)_x, \texttt{next}(\pi)_y, \texttt{rest}(\pi))$ with probability 1.

Plans are found using a noisy variant of the $A^{\star}$ search algorithm \parencite{zhi2020online}.
The standard $A^\star$ algorithm is deterministic, and expands nodes to consider according to the following rule:
$$
n' = \arg \min_{n} d(n) + h(n),
$$
where $d$ is the distance from the start, and $h$ is a heuristic estimate of the distance to the goal.
We use the Manhattan distance as the heuristic function.
To allow $A^\star$ to sample feasible paths probabilistically, we instead expand nodes using the softmax choice rule:
$$
n' \sim \text{softmax}(-(\alpha_d d(n) + \alpha_h h(n))).
$$
Above, $\alpha_d$ and $\alpha_h$ are free parameters that control the degree to which the algorithm greedily or systematically explores the maze.
After search terminates, the optimal path must be reconstructed from the states visited by the search algorithm.
When equally good options are presented for extending the path, we preferentially pick the next state that continues along the straight line trajectory of the past few states.
This straight line bias in $A^\star$ search is intended to capture the comparative ease of pressing the same key over switching keys when moving the agent in the experiment, and produces trajectories that more closely match participant's empirical behavior.

\textbf{Lookahead.}
We assume that the agent running the simulation is performing a one-step lookahead and only checking the validity of the next move against the visual scene. Formally, the lookahead function $\ell$ flags objects only if they intersect the next proposed step of the plan:
$$
\ell(x, y, \pi) = \{o_i \in O | \texttt{next}(\pi) \in o_i \}.
$$
If we find a possible collision, then we add the object to the construal and sample a new plan:
$$
\pi_{\text{new}} \sim A^{\star}(x, y, g, \mathcal{C}_s).
$$

\textbf{Parameter fitting.}
In sum, our model consists of three free parameters $\alpha_d, \alpha_h, \gamma$ that were fit to human data.
We choose these parameters through a grid-search procedure to maximize the Pearson correlation coefficient $r$ between our model's predictions and human responses. Following Ho et al., \parencite*{ho2022people}, separate parameters were fit for each dependent measure in each experiment.
Because the process tracing experiments involve purely planning-based processes and no memory component, we fix $\gamma = 0$ for fitting these experiments and only fit the two $A^\star$ weights $\alpha_d, \alpha_h$.

\subsubsection{Physics}

The physical simulation case is closely analogous to the planning case. 
Here the loop is alternating (1) simulating the motion of the ball using a stochastic physics simulator, and (2) adding an object to the representation and continuing the simulation if an obstacle is observed to be near the simulated path of the ball.
We implement the simplest form of visual lookahead, by assuming that a participant always looks or attends where the simulated ball is in their mind's eye \parencite{gerstenberg2017eye,huber2004ball}, and an object is flagged when it enters their fovea. The state of the simulation $s=(q,v)$ is the current position of the tracked ball $q$ and its velocity $v$, and we initialize the construal to only contain the ball and the goal.

\textbf{Dynamics.} The true state of a plinko board unfolds according to the physical laws of motion, as calculated by the Pymunk 2d physics engine \parencite{blomqvist07}.
However, following work in intuitive physics, we assume that humans lack the ability to exactly simulate the positions and velocities of the objects, and instead instantiate our dynamics using a simulator implementing stochastic variants of Newtonian physics \parencite{smith2013sources}.
Our simulator incorporates three sources of noise \parencite{allen_rapid_2020}. We first apply a random shift to the ball's initial $x$ position, based on a Gaussian distribution centered at the ball with variance $\sigma^2$. When a collision occurs, we randomly rotate the collision normal by drawing a rotation angle from a Von Mises distribution centered at 0 and with variance $\kappa$, and also randomly perturb the collision restitution, by sampling from a truncated normal distribution with variance $s^2$.

\textbf{Lookahead.} We assume that the agent is always visually matching their gaze to the position of the simulated ball, and that visual attention forms a simple circular spotlight centered on the current position of the ball in simulation:
$$\ell_r(q, v) = \{ o_i \in O : ||o_i - (q_x, q_y)|| ^2 \leq r  \}.$$
For all simulations, we fix $r=25$ pixels ($4\%$ of the width of the scenario).

\textbf{Parameter fitting.}
In a supplemental experiment (Section~\ref{sec:supp-param-fit}), participants were asked to rate the likelihood that the ball would hit a randomly chosen object in the world.
Again, we fit simulator noise parameters via grid search: running 500 noisy simulations for each parameter setting, and maximizing the correlation between the collision likelihoods predicted by humans to those predicted by the model. The decay parameter $\gamma$ was fit directly to the recall experiment 2A, again to maximize the correlation coefficient between recall and model predictions; the table of best-fitting parameters is shown in Supplemental table \ref{tab:supp-plinko-fits}. We re-use the exact same fitted parameters to generate model predictions in experiment 2B.

\subsection{Physics experiments details}

\subsubsection{Experiment 2A}

\textbf{Participants.}
We recruited 220 participants through the online platform Prolific. The experiment took 28 minutes on average, and participants were compensated \$6.25 for their time.

\textbf{Procedure.}
Participants engaged in a two stage (prediction and memory probe) task. They first made predictions for where the ball might fall, and were then probed on their memory of the scene they had just made predictions for.
We segmented 1/3rd of the trials into ``critical trials'', and the other 2/3rds into ``filler trials'', with only the critical trials containing the second memory phase.

In the prediction stage, participants viewed the initial scene containing the ball suspended above a series of obstacles, and make predictions about where the ball would land, once let go.
We asked participants to indicate their predictions by clicking 10 times on the ground where they believed the ball would land, and to indicate more or less confidence by concentrating the predictions closer or more diffusely, respectively.

On the critical trials, we presented participants with an additional recall question that probed for their memory of a specific object in the scene.
We generated the memory test by superimposing the previously viewed original scene on top of a distractor scene in which one object had been randomly translated.
The shift ranged from 10 to 50 pixels ($1.7$ to $8.3$ percent of the scene) in magnitude, and was applied in a random direction.
The original and translated object were colored red and blue (randomized between trials); participants were then asked to indicate which object was the original correct one, by indicating their response on a slider ranging from ``definitely sure the red object is in the correct position'', to ``definitely sure the blue object is in the correct position'', with the middle of the slider being labeled ``completely unsure''.
For participants that set the slider to ``completely unsure'', we code their answer as not correct.

Finally, the participants were shown their score for that given trial.
For each prediction, we discretized the ground plane into 10 buckets, and calculated the point score for each prediction $\hat{x}$ and ground truth $x$ as $10 - |\hat{x} - x|$.
We then summed up the point score for each of the 10 predictions to obtain the overall points gained on the trial. This score was not analyzed or used for anything beyond providing participants with motivation.
The points display was followed by a video showing the actual path that the ball would take, when gravity is turned on.

\textbf{Materials.}
Scenarios were constructed to contain a variety of objects that interacted with the ball in different ways.
Critical stimuli contained objects satisfying at least three of four types of ball-object interactions according to noisy physical simulation. \textit{Early collision} objects were hit by the ball with $> 95\%$ probability, and were always hit in the top half of the environment, \textit{late collision} objects were also hit with $> 95\%$ probability but were contacted in the lower half of the world, \textit{maybe collision} objects were hit by the ball between $40$ and $60$ percent of the time, and \textit{no collision} objects were never hit by the ball. We used the noisy simulator from Allen et al., \parencite*{allen_rapid_2020} to estimate these probabilities for each world, as stimulus creation necessarily was done before experiments and parameter fitting.
After running Experiment 2A, a mixed effects model regressing object type on memory with random intercepts by participants showed a statistically significant effect ($\chi^2(3)=268.37~p\approx 0$), verifying that our procedure did produce objects with substantially different memory profiles.
The trials used in this experiment can be seen in the heatmaps of Extended Figures \ref{fig:extended-plinko-heatmaps1} and \ref{fig:extended-plinko-heatmaps2}, and code for running the experiments can be found at \url{https://github.com/chentoast/physics_repr}.

Filler trials were generated by procedurally filling the world with 8-12 objects, and placing the ball at a random horizontal position. We then manually narrowed down the list of generated candidates to 12 worlds.






\subsubsection{Experiment 2B: Distinguishing \shortmodel~and VGC}

\textbf{Participants.}

We recruited 50 participants from Prolific. The experiment took 15 minutes on average, and participants were compensated \$4.50 for their time.

\textbf{Procedure.}
The procedure was identical to Experiment 1 except that the predictions were discretized: the ground area was segmented into five ``buckets'' and participants were instructed to select the bucket that the ball was most likely to fall into. The feedback and memory test phases were otherwise the same.

\textbf{Materials.}
The filler trials were the same as in Experiment 1. However, the critical trials were constructed to have two different types of obstacles for the memory test. These obstacle types were chosen because the \shortmodel~and VGC models would produce a different ordering of representational strength between the two types.

Half of the trials were constructed to have a ``counterfactually irrelevant'' object: an obstacle that was certain to be hit by the ball (similar to the ``early collision'' obstacles), but in such a way that, due to the geometry of other obstacles, the bucket that the ball was almost certain to end up in ($>95\%$ under the simulation model) was the same bucket that the ball would end up in if the lure obstacle were removed from the scene. These objects had high relevance under the \shortmodel~model as they were always contacted, but because the outcome would barely differ, had little relevance under the VGC model.

The other half of the trials were constructed to have a ``counterfactually relevant'' object: an obstacle that was only likely to be contacted about half the time according to the noisy simulation model, but if it is contacted, would certainly change the ending bucket the ball would land in as compared to if that obstacle were not there. These objects would have high relevance under the VGC model because its inclusion always makes a difference in the outcome, but only moderate relevance under the \shortmodel~model as they are included only in the half of simulations that involves the ball contacting them.

The trials used in this experiment can be seen in the heatmaps of Extended Figure \ref{fig:extended-plinko-vgc-heatmaps1},\ref{fig:extended-plinko-vgc-heatmaps2} and found at \url{https://github.com/chentoast/physics_repr}



\section{Author Contributions}
TC, SC, KS, and JT formulated the model; TC implemented the model, ran the behavioral experiments, and analyzed the data; TC, KA, and KS conceived of the main project; TC, SC, KS wrote the manuscript with assistance from KA and JT.


\printbibliography
\newpage

\setcounter{secnumdepth}{0}
\section{Extended Data Figures}
\setcounter{secnumdepth}{3}

\renewcommand{\thetable}{E\arabic{table}}
\renewcommand{\thefigure}{E\arabic{figure}}
\renewcommand{\theequation}{E\arabic{equation}}
\setcounter{table}{0}
\setcounter{figure}{0}
\setcounter{footnote}{0}
\setcounter{equation}{0}

\begin{figure}
    \centering
    \includegraphics[width=0.70\linewidth]{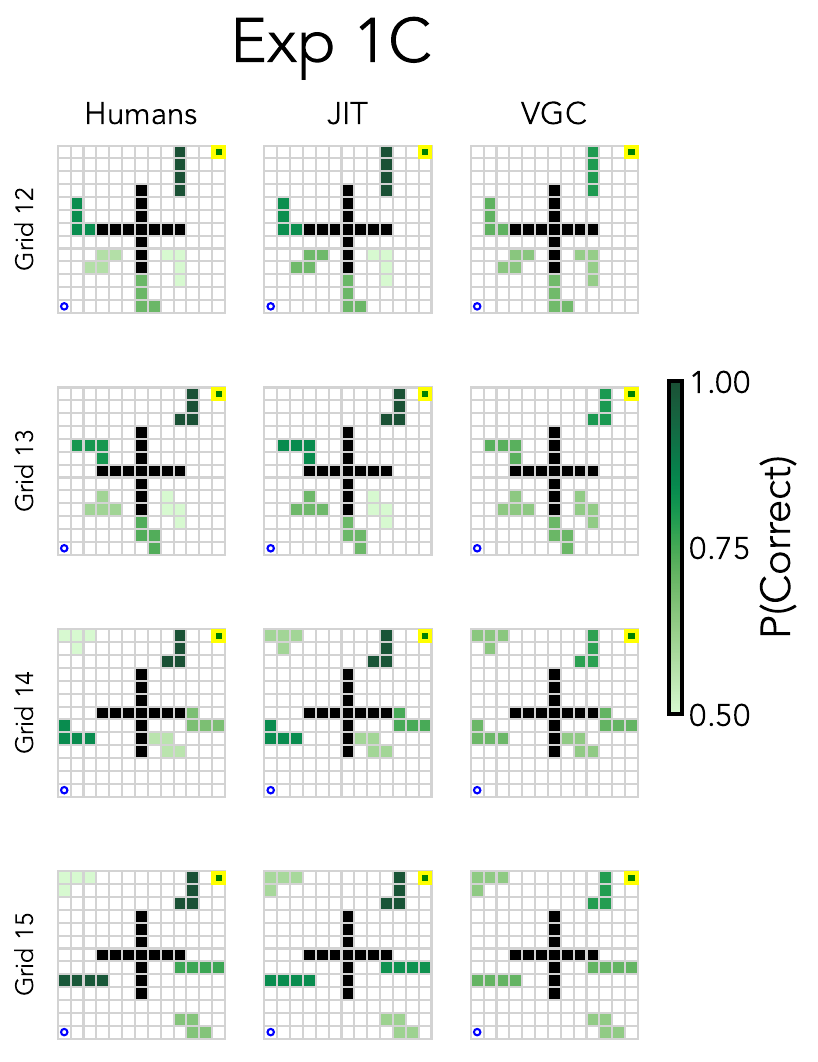}
    \caption{Model predictions and human results for the four mazes tested in Experiment 1C. The center cross obstacle was present in all mazes, and was not probed in the memory tests. Darker green objects were more likely to be recalled by participants, and more likely to be represented in the models.}
    \label{fig:extended-navigation-recallheatmap}
\end{figure}

\begin{figure}
    \centering
    \includegraphics[width=0.83\linewidth]{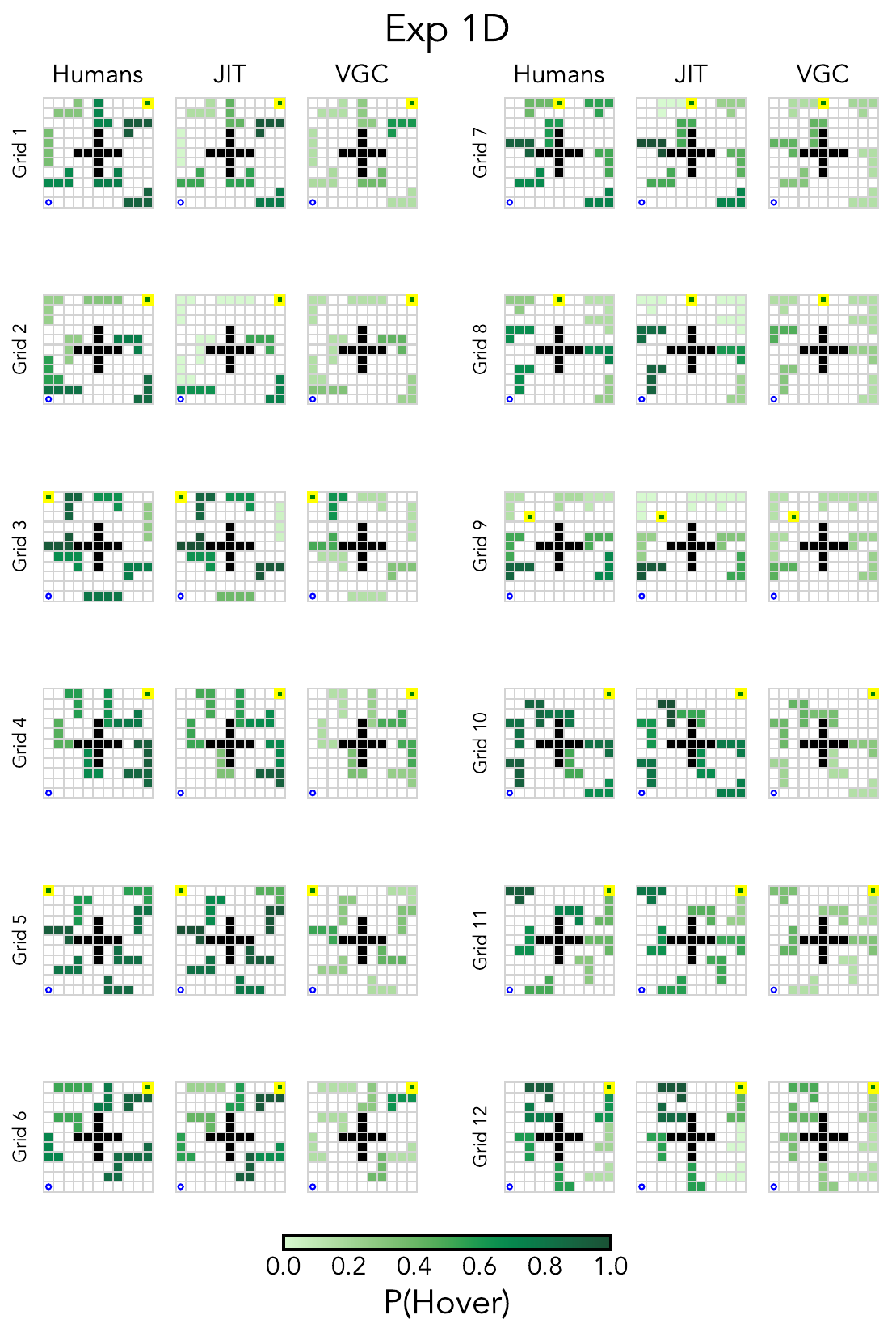}
    \caption{Model and human predictions for the mazes tested in Experiment 1D. Darker green objects were more likely to be hovered by the participant's mouse while planning, and more likely to be represented in the models.}
    \label{fig:extended-navigation-hoverheatmap1}
\end{figure}



\begin{figure}
    \centering
    \includegraphics[width=0.70\linewidth]{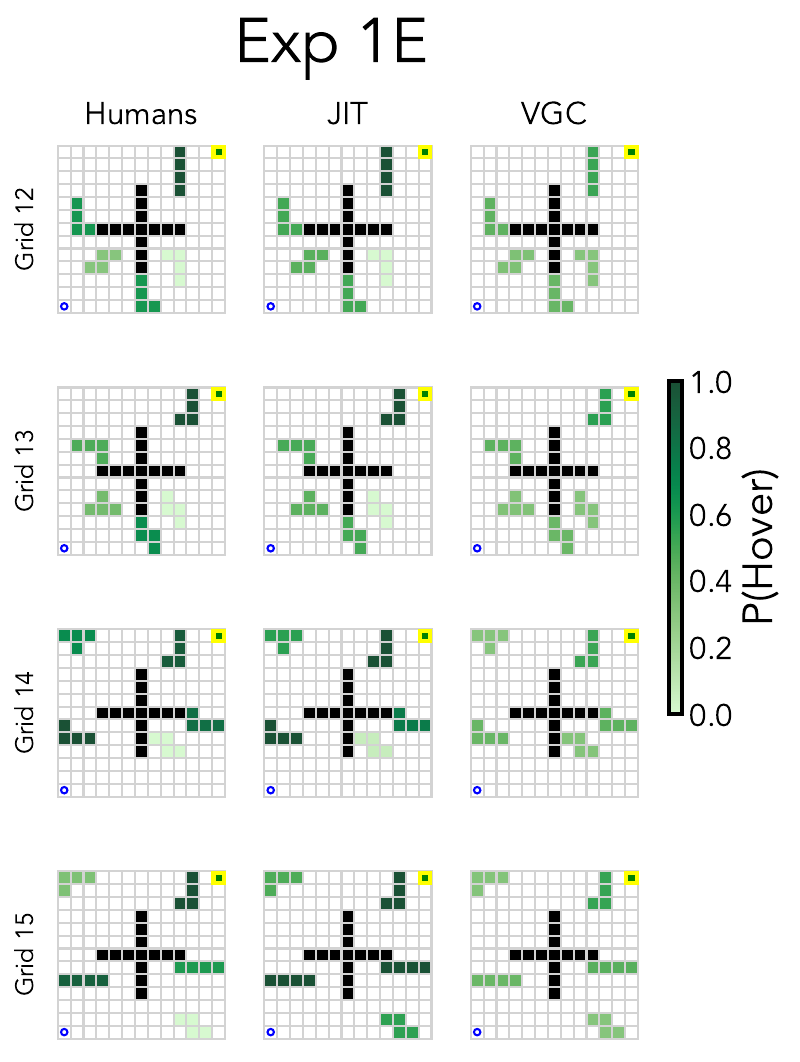}
    \caption{Model and human predictions for the four mazes tested in Experiment 1E. Darker green objects were more likely to be hovered by the participant's mouse while planning, and more likely to be represented in the models.}
    \label{fig:extended-navigation-hoverheatmap4}
\end{figure}

\begin{figure}
    \centering
    \includegraphics[width=0.99\linewidth]{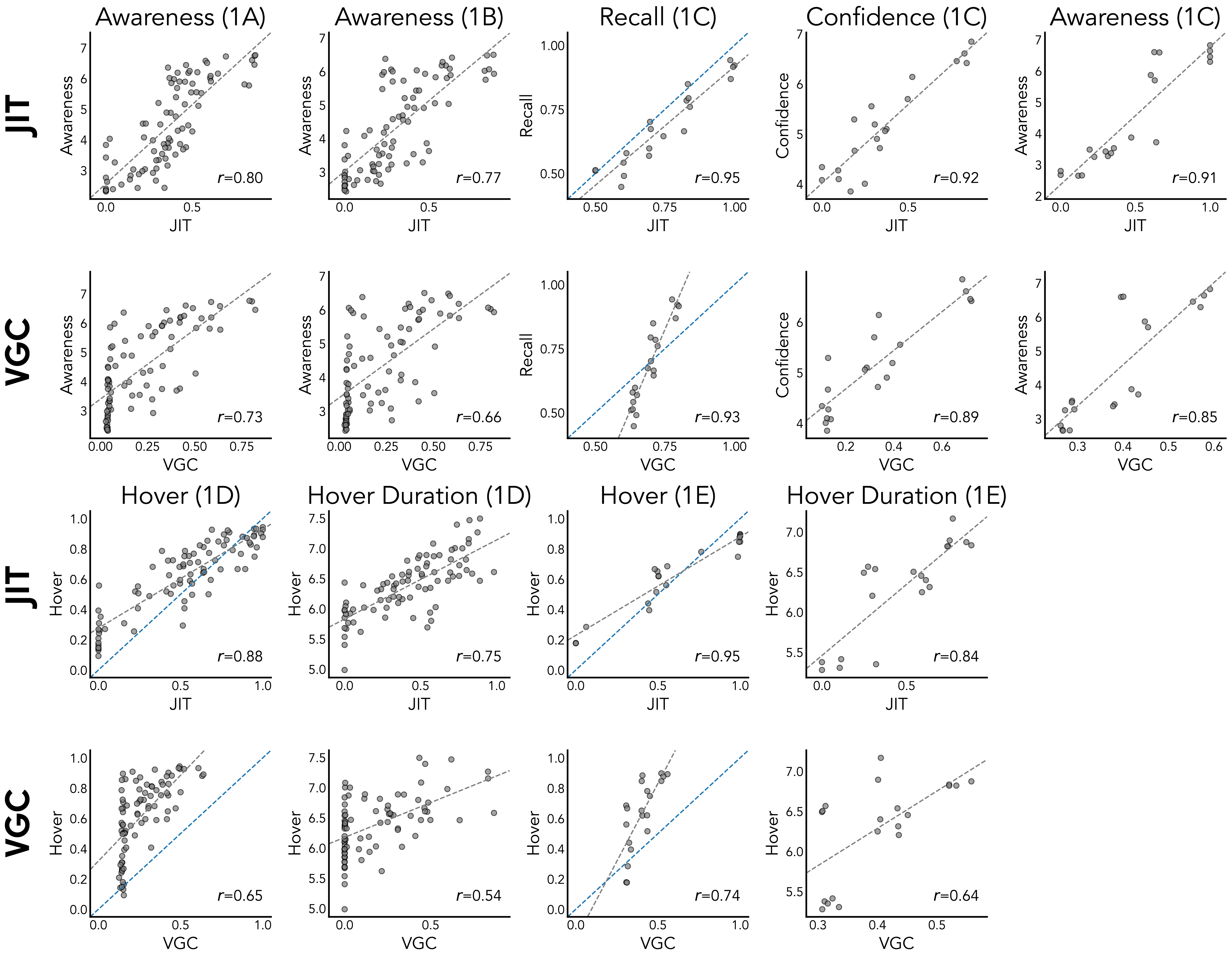}
    \caption{Scatterplots comparing JIT (upper rows) and VGC (lower rows) predictions against human results for each of the 9 dependent measures studied in \textcite{ho2022people}. Awareness denotes explicit judgments of how aware participants were of each object in the maze; Recall denotes the probability of correct recall in the memory probe; Confidence denotes participants' confidence in their answer on the memory probe; Hover denotes the probability an object is hovered by the mouse during the process-tracing experiments; and Hover Duration is the average log duration that each object was hovered over for.
    Gray dashed lines denote the line of best fit; for measures that lie in a $[0, 1]$ range, the blue dashed line denotes the identity line.
    }
    \label{fig:extended-navigation-scatterplots}
\end{figure}

\begin{figure}
    \centering
    \includegraphics[width=0.99\linewidth]{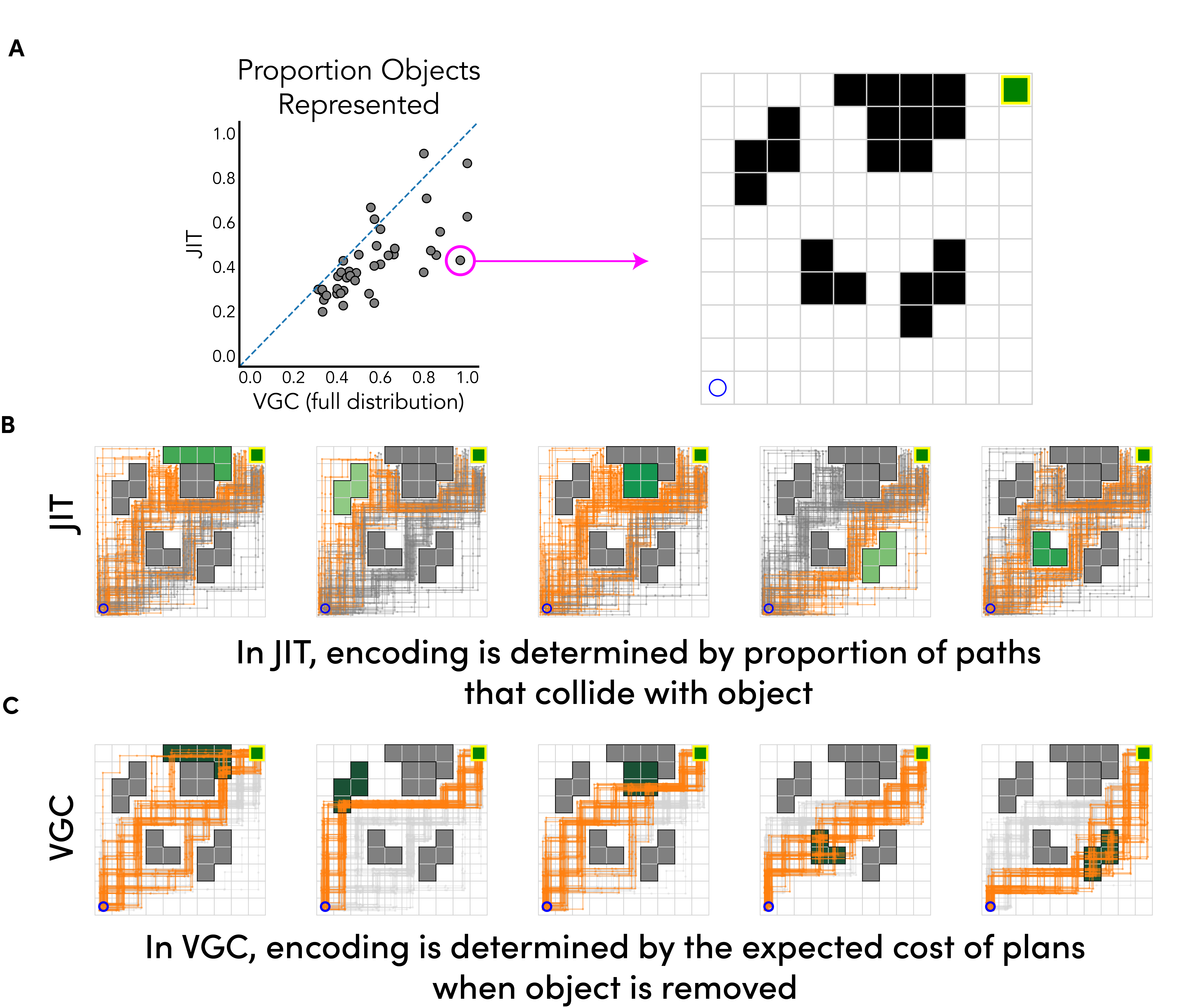}
    \caption{\textbf{A.} A procedurally generated world where JIT represents less than half the objects as VGC. The scatterplot is the same as Fig.~\ref{fig:navigation-results}C, labeling the world in current focus.
    \textbf{B.} JIT predictions for each object, with darker green representing objects more likely to be included in the construal. Gray lines denote trajectories that did not encounter the highlighted object, and orange lines denote trajectories that did encounter and represent the object.
    \textbf{C.} VGC predictions for each object, with darker green representing higher utility objects (note that all objects have high utility).
    Lines denote rollouts sampled from the construal policy that represents all objects except for the highlighted one.
    Gray trajectories attain the optimal trajectory value, while orange lines do not.
    For objects on the periphery of the maze, JIT samples paths that often avoid those objects entirely, whereas when those same objects are excluded from the construal, VGC finds many paths that will consistently try to move through the excluded objects and thus incur low value.
    }
    \label{fig:extended-theory-example}
\end{figure}

\begin{figure}[p]
    \centering
    \includegraphics[width=0.90\linewidth]{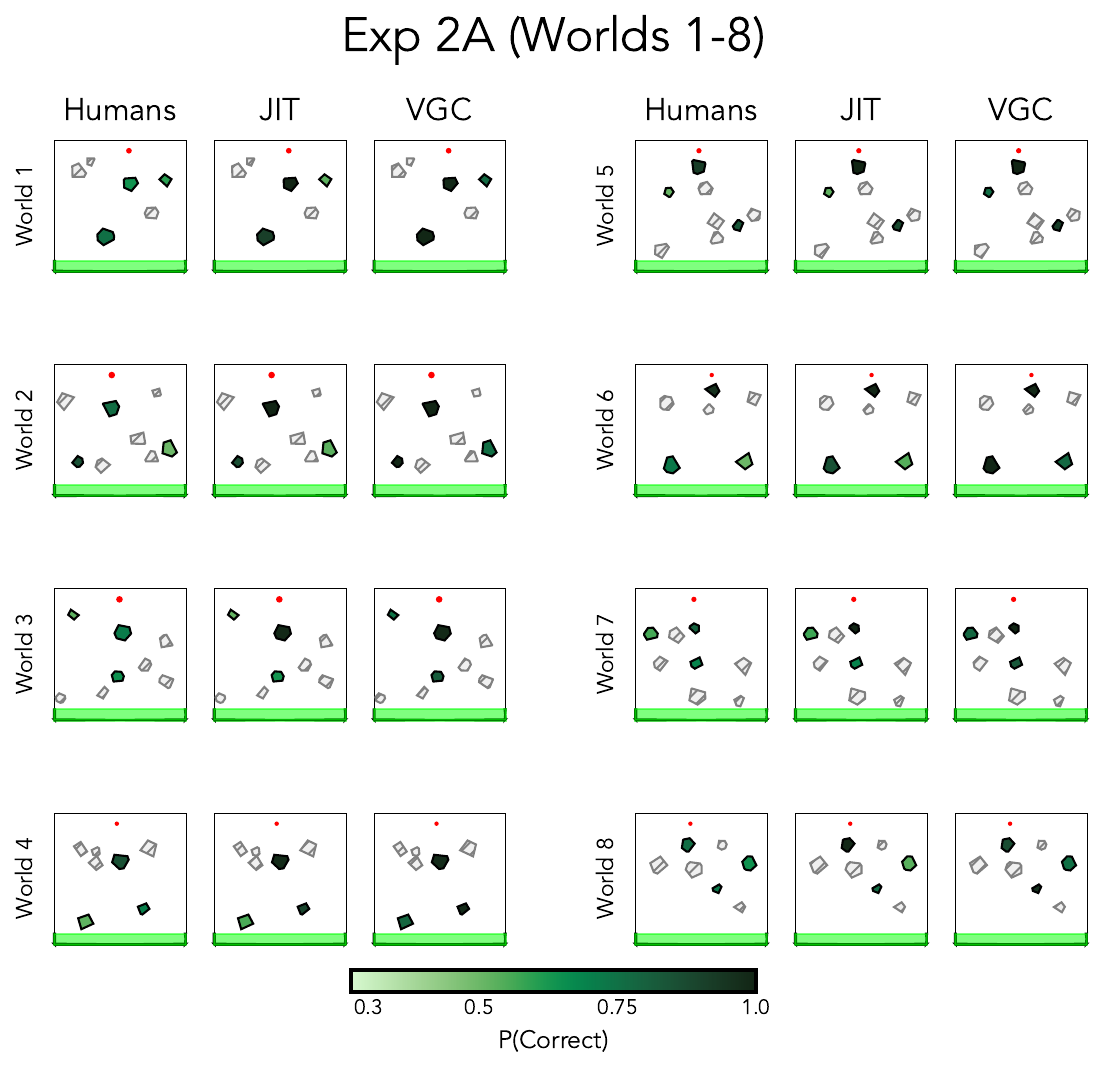}
    \caption{Model predictions and human recall for the worlds 1-8 in Experiment 2A. Darker green denotes higher recall accuracy, and lighter green denotes lower recall accuracy. We did not probe memory for the light gray hatched objects.}
    \label{fig:extended-plinko-heatmaps1}
\end{figure}

\begin{figure}
    \centering
    \includegraphics[width=0.90\linewidth]{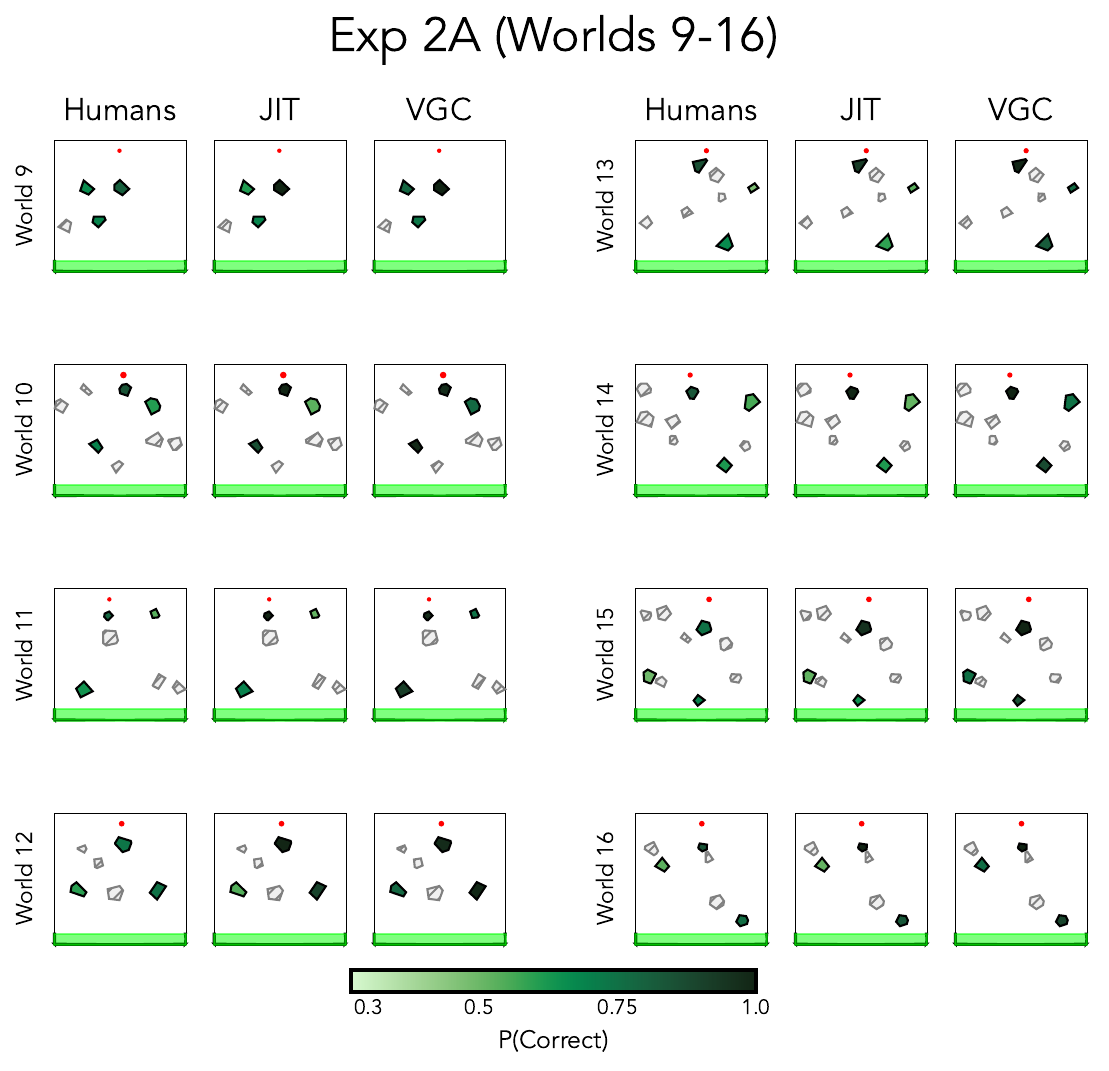}
    \caption{Model predictions and human recall for the worlds 9-16 in Experiment 2A. Darker green denotes higher recall accuracy, and lighter green denotes lower recall accuracy. We did not probe memory for the light gray hatched objects.}
    \label{fig:extended-plinko-heatmaps2}
\end{figure}


\begin{figure}
    \centering
    \includegraphics[width=0.99\linewidth]{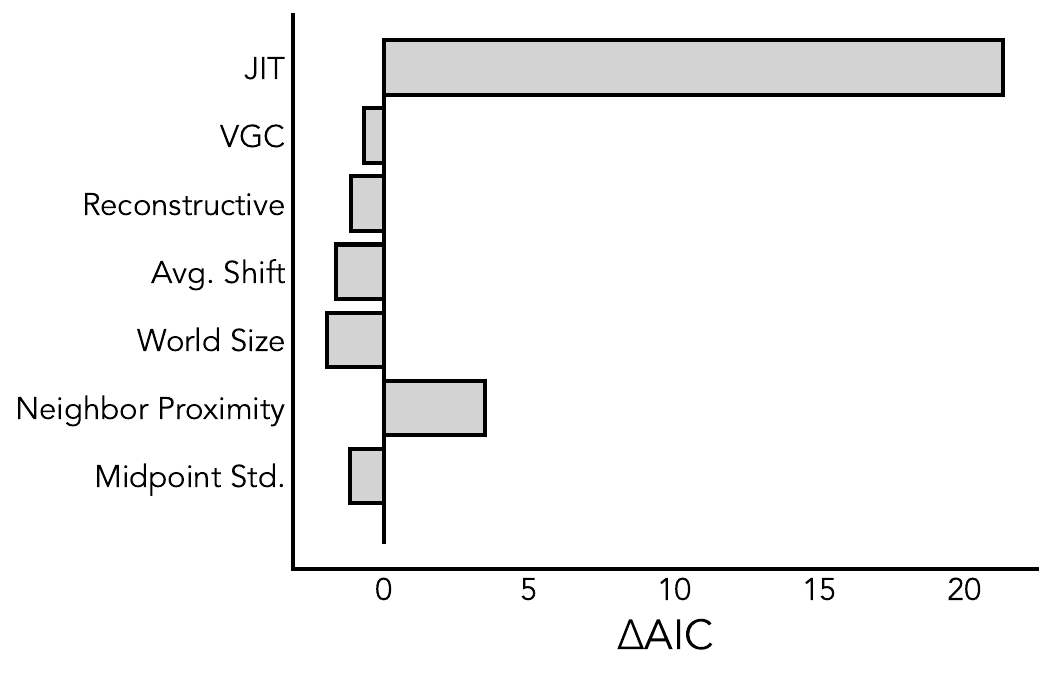}
    \caption{Model comparisons visualizing predictive power of the JIT model and a set of alternative predictors for human memory in Experiment 2A. In order from top to bottom, these alternative predictors are: VGC, a \textit{reconstructive baseline} (see Supplemental Fig.~\ref{fig:supp-metacognitive-results}) using counterfactual simulation to determine recall, the average distance of the distractor stimulus to the true object, number of objects in the world, proximity of the probed object to its nearest neighboring object, and the average standard deviation of positions. Bar lengths denote the change in AIC when removing that predictor from a mixed effects model with random intercepts by participant and all predictors as covariates. Except for the neighbor proximity, all other predictors do not explain substantial variance on top of JIT.}
    \label{fig:extended-plinko-modelcomparison}
\end{figure}

\begin{figure}
    \centering
    \includegraphics[width=0.95\linewidth]{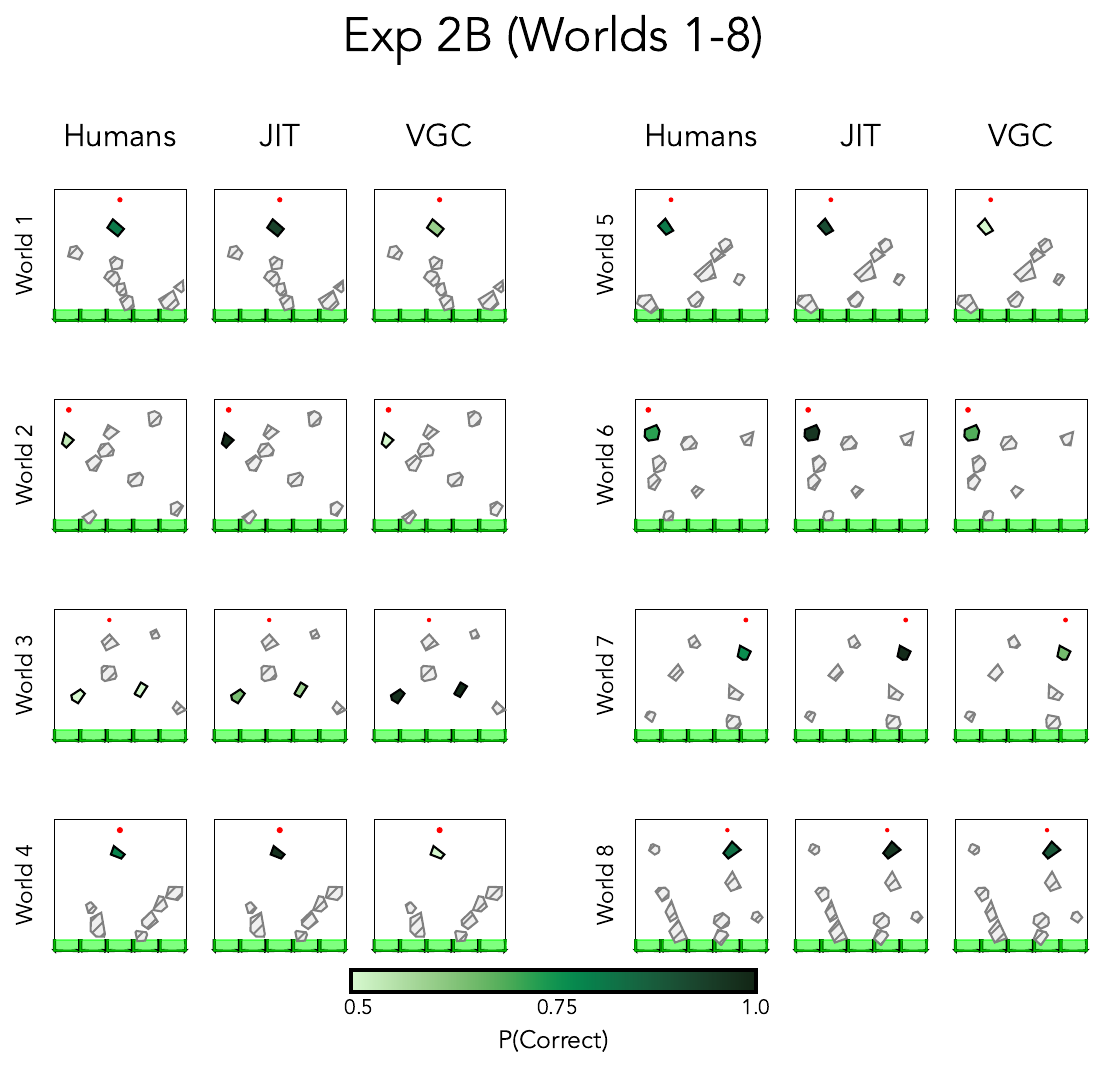}
    \caption{Model predictions and human recall for the worlds 1-8 in Experiment 2B. Darker green denotes higher recall accuracy, and lighter green denotes lower recall accuracy. We did not probe memory for the light gray hatched objects.}
    \label{fig:extended-plinko-vgc-heatmaps1}
\end{figure}

\begin{figure}
    \centering
    \includegraphics[width=0.95\linewidth]{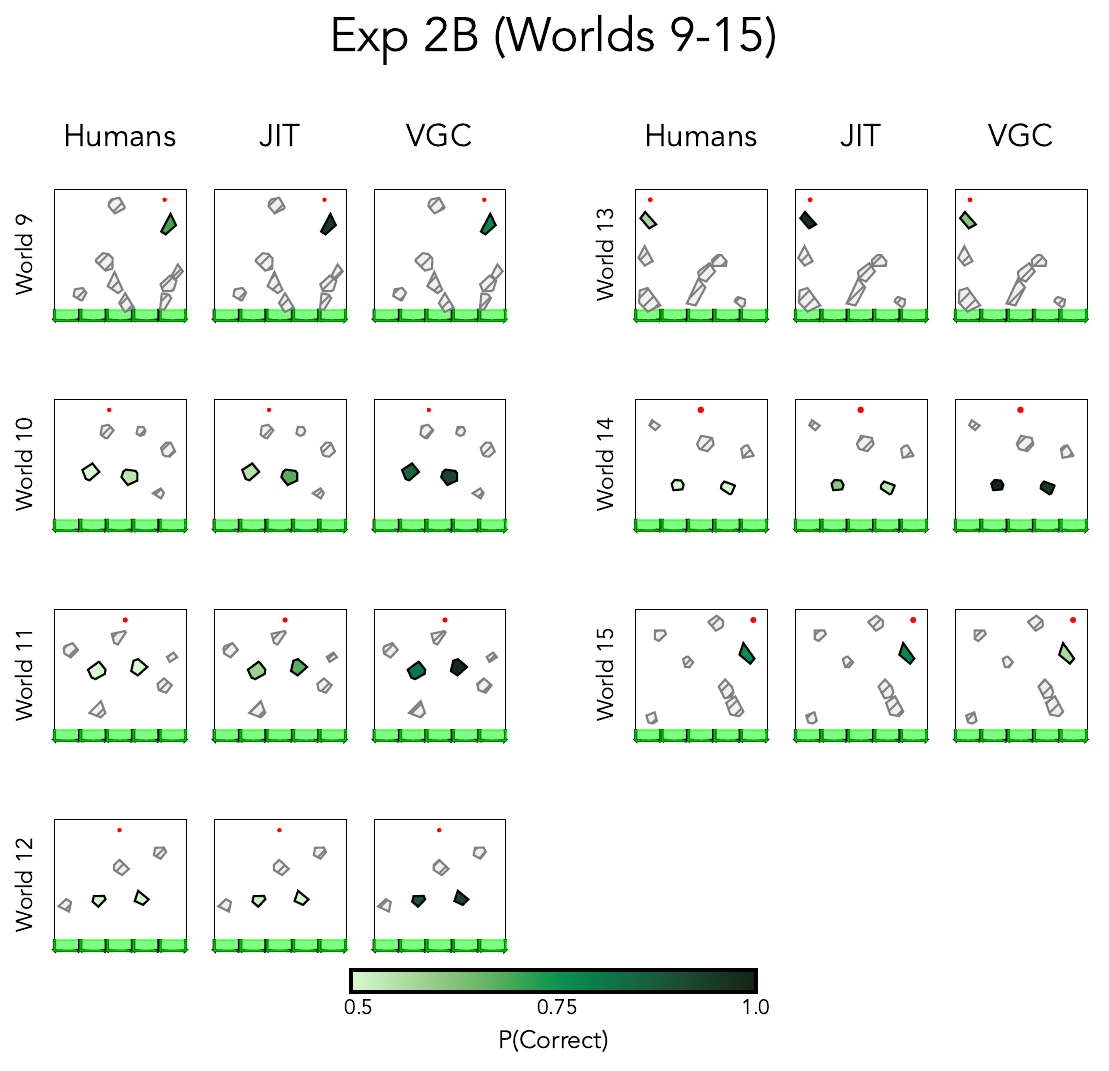}
    \caption{Model predictions and human recall for the worlds 9-15 in Experiment 2B. Darker green denotes higher recall accuracy, and lighter green denotes lower recall accuracy. We did not probe memory for the light gray hatched objects.}
    \label{fig:extended-plinko-vgc-heatmaps2}
\end{figure}


\newpage

\setcounter{secnumdepth}{0}
\section{Supplementary Materials}
\setcounter{secnumdepth}{3}

\renewcommand{\thesection}{S\arabic{section}}
\renewcommand{\thetable}{S\arabic{table}}
\renewcommand{\thefigure}{S\arabic{figure}}
\renewcommand{\theequation}{S\arabic{equation}}
\setcounter{section}{0}
\setcounter{table}{0}
\setcounter{figure}{0}
\setcounter{footnote}{0}
\setcounter{equation}{0}



\section{Additional physics experiments}


We ran two additional experiments to test whether recall on the physics memory probe tasks might have been driven by shallow visual features instead of the depth of representational encoding. In Experiment S1, we change the nature of some obstacles from ``foreground'' to ``background'' to test whether memory is driven by simply looking at the object. In Experiment S2, we add ``teleporters'' with different functionality for different sets of participants to test whether it is simple visual layout of the scene driving recall. In both cases, we find that encoding relies on whether objects are relevant to the simulation of the ball's motion, and not simply visual features.

Additionally, we ran Experiment S3, asking participants to directly rate the probability critical blocks would be contacted, in order to calibrate parameters for the simulation model used in Experiments 2A and 2B.

\subsection{Experiment S1: Testing for visual attention}
\label{sec:supp-attention}

\subsubsection{Participants}

We recruited 100 participants using Prolific. The median time to complete the task was 22 minutes, and participants were compensated \$5.00.

\subsubsection{Procedure}

The procedure was identical to Physics Experiment 2A in the main paper. 
We used the same set of 16 critical worlds, but additionally separated objects into two categories, colored either brown or blue.
Foreground objects are solid, and the ball bounces off of them as normal, whereas background objects are completely ephemeral, and the ball passes through them as if they were painted sections of the wall.
The color used to indicate object category was counterbalanced across participants.

We also re-use the same memory probes as Experiment 2A. Because colors denote the solidity of objects in the task, instead of marking the target and lure objects by different colors, we instead display query objects as unfilled outlines, and named them either object A or object B. Participants were then asked to indicate whether object A or B was in the correct position.



\subsubsection{Results}

Results are shown in Supplemental Figure \ref{fig:supp-fgbg}.
We again discretized objects into the four object categories defined in the Methods section.
A mixed effects logistic regression with random interce2pts by participant found a strong main effect of object category, with memory for foreground objects being significantly higher than background objects ($\chi^2(1) = 7.19,~p=0.007$).
Paired t-tests show that this effect was most significant in the early collision objects: $t(15) = 4.23,~p=0.007$, and non-existent in the no-collision objects: $t(15)=-0.37,~p=0.71$.
Effects for the middle two object categories are weaker but directionally correct; we find a marginally significant effect for maybe collisions: $t(7) = 2.52,~p=0.04$, and no effect for late collisions: $t(7) = .78,~p=0.46.$

\begin{figure}
    \centering
    \includegraphics[width=0.49\linewidth]{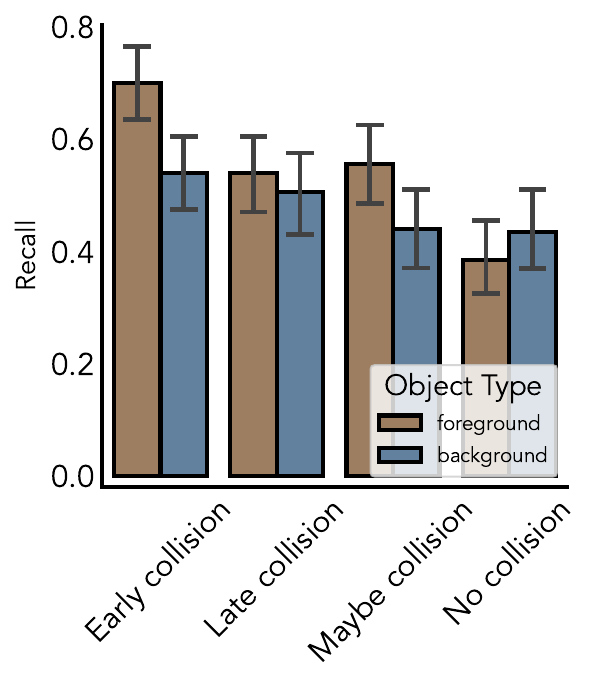}
    \caption{
        Human recall for the foreground and background variants of the same object. Objects were binned into one of four categories: collision early (contacted $>95\%$ of the time under noisy simulation, and located in the top half of the environment), collision late ($>95\%$ collision probability, located in lower half of environment), maybe collision (between $40$ and $60$ percent collision probability), and no collision objects.
        Memory was worse for objects presented in background form; this effect was modulated by object condition, such that foreground objects clearly hit by the ball show a greater difference in memory from objects not hit by the ball at all.
    }
    \label{fig:supp-fgbg}
\end{figure}

\subsection{Experiment S2: Testing for purely visual processes}
\label{sec:supp-visual}

While experiment S1 shows that memory for objects cannot be entirely explained by visual attention during simulation, representations could still be driven by a primarily visual process as opposed to simulation.
In Experiment S2, we present participants with matched scenes that are almost visually identical, yet have very different simulation profiles, and show that human memory patterns differ substantially between these pairs.

\subsubsection{Participants}

We recruited 100 participants on Prolific. The median time spent on the task was 18 minutes and participants were compensated \$5.25 for their time.

\subsubsection{Procedure}

We generate a new set of worlds containing teleporter objects.
A teleporter pair consists of an entrance and an exit; when the ball contacts the entrance, it will instantly re-appear at the center of the exit with the same velocity.
The teleporter exit is not a solid object, and objects will simply pass through it.
Just as in Experiment S1, the teleporter entry and exit objects were labeled by unique colors that were counterbalanced across participants.
Matched scenes were generated by swapping the positions of the teleporter entry and exit. In ``entry'' scenes, the ball contacts the teleporter entrance and re-appears at the exit, and in ``exit'' scenes, the ball instead touches the teleporter exit and thus falls through uninterrupted.

\subsubsection{Materials}

We generated an initial set of candidate worlds using procedural generation procedure.
After initializing a world with the ball and a set of 5-10 objects, we place the teleporters at random free positions, with the constraint that either the entrance or the exit must be the first object touched by the ball.
We then proceeded to filter down the candidate set into a set of 10 worlds, and manually edited them to more cleanly distinguish outcomes in the ``entry'' versus ``exit'' variants.

\subsubsection{Results}

Results are shown in supplemental figure \ref{fig:supp-teleport}.
A paired t-test shows that memory for the same object differs significantly depending on whether the object was hit by the ball or not: ($t(17)=4.07,~p=0.008$).
More specifically, we find that while memory for objects that are hit are recalled better than chance: mean: $0.69$, $t(17)=7.79,~p\approx0$, memory for objects that were not hit by the ball was not higher than chance. This is true both for objects that were not hit because of the configuration of the teleporters: $t(17)=1.45,~p=0.17$, and for objects located in the periphery, far from the ball's trajectory in either condition: $t(8)=-0.64,~p=0.54$.

\begin{figure}
    \centering
    \includegraphics[width=0.49\linewidth]{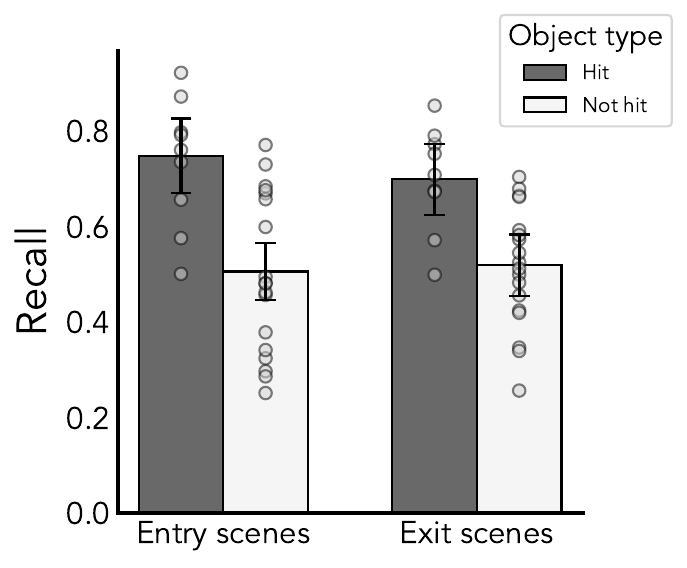}
    \includegraphics[width=0.49\linewidth]{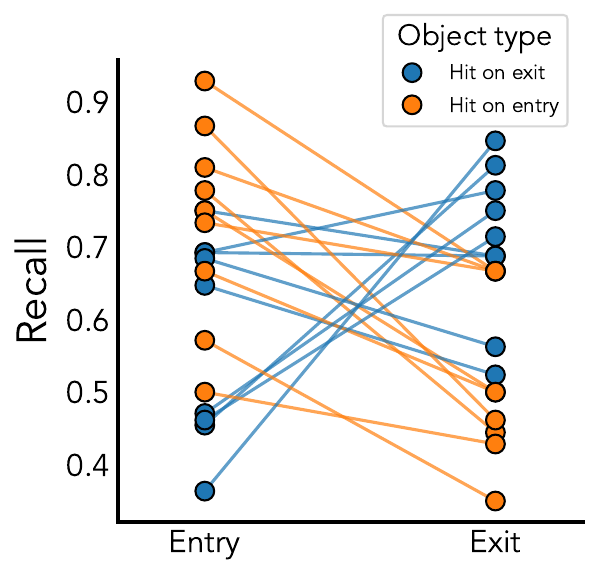}
    \caption{Left: memory for objects that collide with the ball (blue) and objects that do not (orange) in each type of world. Right: differences in memory for each object depending on whether participants viewed that object in an entry world or exit world. Blue objects are hit on exit worlds but not hit on entry worlds, and vice versa for orange objects. Lines connect paired objects.}
    \label{fig:supp-teleport}
\end{figure}

\subsection{Experiment S3: Measuring human simulations}
\label{sec:supp-param-fit}

To fit the noise parameters of the simulator, we run an additional study where we directly ask participants for the likelihood that the ball will hit a given object. We then fit the noise parameters to maximize the Pearson correlation coefficient between the model predicted probability of the ball hitting an object with the empirical human predictions.

\subsubsection{Participants}

We recruited 50 participants on Prolific. The median time to perform the task was 13 minutes and participants were compensated \$3.50 for their time.

\subsubsection{Procedure}

We use the same set of 16 critical worlds as Experiment 2A. 
After making predictions for where the ball would land, participants were taken to the next stage, where they were shown the same world with one object highlighted, and asked how likely they thought the ball was to hit that object.
Participants gave their responses on a slider from ``Definitely not going to hit'', to ``Definitely going to hit''.

\subsubsection{Results}
The fitted parameter values are displayed in Table \ref{tab:supp-plinko-fits}. Noisy simulations with the fitted parameter values are able to closely match the human-normed collision likelihoods for each object $(r=0.97, \text{rmse}=0.09)$, as shown in Supplemental Figure~\ref{fig:supp-plinko-collision}.

\begin{figure}
    \centering
    \includegraphics[width=0.5\linewidth]{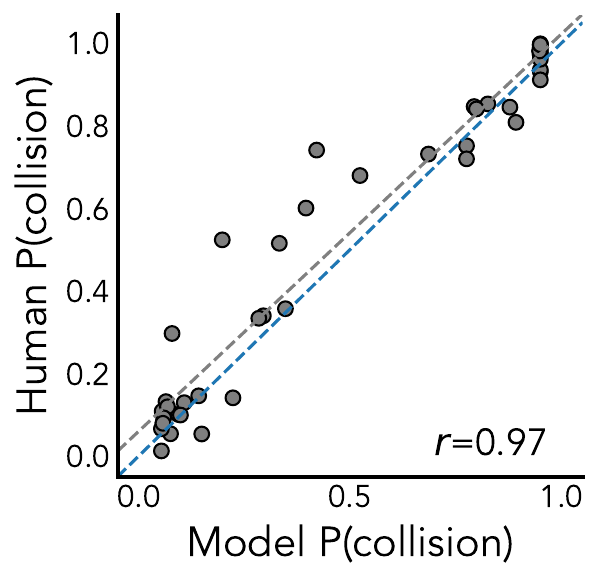}
    \caption{Model predicted probabilities that the ball will collide with a target object compared to human predicted probabilities. Model predictions were generated using the simulation parameters fit in Supplemental Experiment 3.}
    \label{fig:supp-plinko-collision}
\end{figure}

\begin{table}[t]
    \centering
    \begin{tabular}{ccc}
    \toprule
    $\sigma^2$ & $\kappa$ & $s^2$ \\
    \midrule
    5.0 & 0.8 & 0.6 \\
    \bottomrule
    \end{tabular}
    \qquad\qquad\qquad
    \begin{tabular}{l c}
    \toprule
     & parameter \\
    \midrule
    JIT & $\gamma=0.0$ \\
    VGC & $\alpha=20.0$ \\
    \bottomrule
    \end{tabular}
    \caption{Left: best-fitting simulation noise parameters for the physics experiments, fit to match human predictions in supplemental experiment S3. Right: other model parameters (the decay parameter $\gamma$ for JIT, and softmax temperature $\alpha$ for VGC) were fit to maximize correlation with human recall in Experiment 2A. These same parameters were re-used to generate model predictions for experiment 2B.}
    \label{tab:supp-plinko-fits}
\end{table}

\section{Additional analyses and results}

\subsection{Details on the planning efficiency simulation analysis}

Here, we provide additional details on the simulation analyses visualized in Main text Fig. \ref{fig:navigation-results}.

We procedurally generate a set of 40 10x10 grid worlds to serve as our evaluation set. Grid worlds were constrained to contain between 5-10 objects and contain at least one object-free path from start to goal. Each object was made up of 3-8 continguous tiles.
When comparing JIT and VGC, we used fixed-parameter versions of both models. For JIT, we used $\alpha_d=0, \alpha_h=1,$ turning $A^\star$ search into a stochastic greedy best-first search, and fixed $\gamma=0$. For VGC, we use the static model variant, and set $\alpha = 0.1$, mirroring the parameter setting in \textcite{ho2022people}.

Extended data Figure \ref{fig:extended-theory-example} shows a case study of a world with extreme disparity in the amount of objects that JIT and VGC represent. In this world, the maximally valuable construal chooses to represent every object, while JIT only represents only half of the objects on average.
This discrepancy can be explained by visual inspection of the optimized policies and plans found by both models.
The \shortmodel~model samples trajectories that mostly miss peripheral objects on the edges of the maze, resulting in low encoding strength for those objects.
Conversely, for VGC, those same objects cause catastrophic failure modes for the optimal construal policy when removed; thus necessitating their inclusion. For example, when removing object 2 from the construal (second column of Extended Fig. \ref{fig:extended-theory-example}), plans that begin by moving upwards may become permanently stuck as they repeatedly attempt to move right, directly into the object. These plans have a value of $-\infty$ as they never reach the goal, and thus dramatically deflate the utility of any construal that does not include object 2.

%
\subsection{Details on VGC implementation for physics tasks}
\label{sec:supp-vgc-phys}

The value-guided construal model aims to find a construal that maximizes the following \textit{value of representation} (VOR) objective:
$$
\text{VOR}(c) = U(c) - C(c),
$$
where $U(c)$ denotes the utility of simulating or planning with the construal, and $C(c)$ denotes the representational cost of the construal. As in the planning domain, we take the cost of representation to be the size of the construal: $C(c) = |c|$. The utility of simulation is given by the counterfactual difference (as measured by the 1-Wasserstein distance) between predictions made by noisy simulation under the construal $c$ and \textit{ground truth predictions} $q_\text{true}$ obtained by noisy simulation with every object included:
$$U(c) = W_1(q_s, q_{\text{true}}).$$

\begin{figure}
    \centering
    \includegraphics[width=0.48\linewidth]{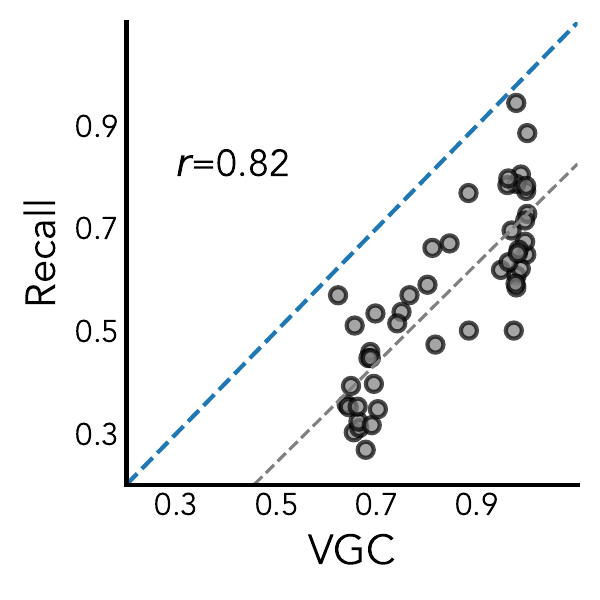}
    \includegraphics[width=0.48\linewidth]{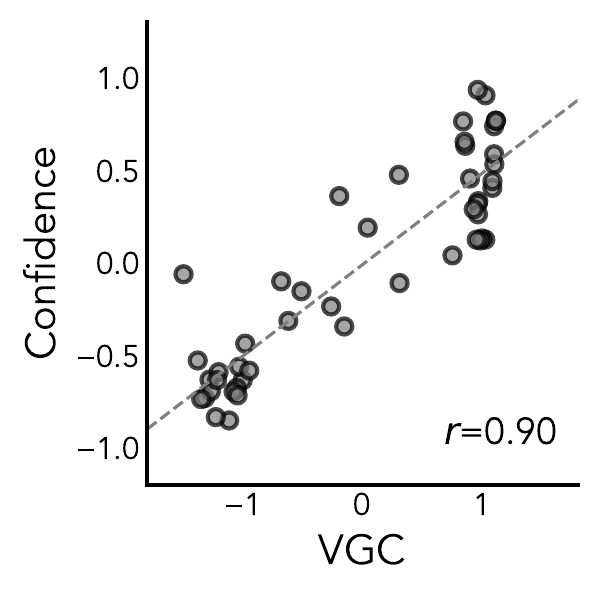}
    \caption{Scatterplots showing predictions from the physics VGC model against Recall (left) and Z-scored Confidence (right) in physics Experiment 2A.}
    \label{fig:supp-vgc-plinko}
\end{figure}

Instead of selecting the maximally valuable construal, we instead follow \textcite{ho2022people} and assume decision makers probabilistically choose a construal via the Luce choice rule, with a temperature parameter $\alpha$:
$$
p(c) \propto e^{\text{VOR}(c) / \alpha}
$$
We obtain predictions for each object by marginalizing over $p(c)$. For simulation, we use the same set of noise parameters fit to supplementary experiment 3 (see Supplementary Table~\ref{tab:supp-plinko-fits}, and we fit the softmax temperature $\alpha$ via grid-search to maximize the Pearson correlation coefficient $r$ to human recall. The best-fitting softmax temperature was $\alpha=20.03$; results are shown in Fig.~\ref{fig:supp-vgc-plinko}

\begin{figure}
    \centering
    \includegraphics[width=0.49\linewidth]{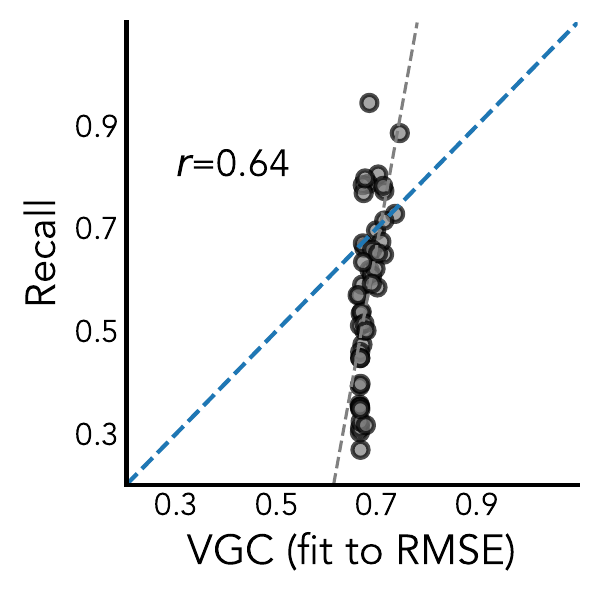}
    \includegraphics[width=0.49\linewidth]{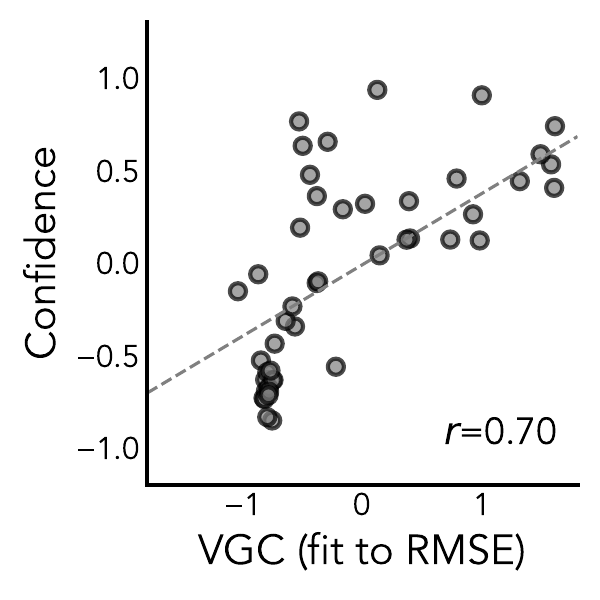}
    \caption{Results from fitting VGC parameters to minimize RMSE instead of maximize correlation. Scatterplots show VGC predictions against human recall (left), and confidence in recall (right).}
    \label{fig:supp-vgc-plinko-rmse}
\end{figure}

We also alternatively explored fitting VGC to minimize RMSE instead of maximizing correlation. The best fitting softmax temperature was $\alpha=370.09$, and the results are shown in Fig.~\ref{fig:supp-vgc-plinko-rmse}.

In experiment 2B, the outcome of the ball is discretized into five uniformly spaced bins $b_1, \ldots b_5$. This necessitates a change to the utility function of VGC, as the wasserstein distance is only defined on metric spaces and hence does not apply to categorical distributions. Instead, we use the total variation (TV) distance between the distribution of bin counts as our utility, defined as half the summed absolute difference between the probability mass functions of the two distributions:
$$
U(c) = TV(q_x, q_{true}) = \frac{1}{2}\sum_{1 \leq i \leq 5} |q_x(b_i) - q_{true}(b_i)|.
$$
Because the temperature parameter $\alpha$ and construal cost $C(c)$ are scaled relative to the utility function in 2A, we multiply the 2B utility function by a fixed constant, to ensure that the average utility of a construal in 2B is equal to the average utility of a construal in 2A.

\subsection{Alternative Decision Rules}
For all physics experiments, we used a simple decision rule in which the probability of responding correctly to the memory probe was linearly proportional to the average probability that the object was represented in working memory.
Here, we test an alternative based on principles of signal detection theory, but find that the simple decision rule describes participants' behavior better.

We represent uncertainty due to memory imprecision about an object's position as a 2D gaussian distribution.
To formalize how encoding an object affects uncertainty, we place a maximally diffuse prior over object position: $p(o) \propto 1$, and model encoding an object as observing a sample $x_i$ from a distribution $N(o_{true}, \kappa)$. We assume that participants always observe an unbiased sample of the true object's position: $x_i = o_{true}$.
The updated uncertainty about the true position of the object after encoding an object $n$ times is then given by Bayes' rule:
$$
p(o| x_1, \ldots x_n) \propto p(x_1, \ldots x_n) p(o) = \text{Normal}(o_{true}, \bar\kappa),
$$
where $\bar\kappa = \sqrt{\dfrac{\kappa^2}{n}}.$
At test time, we evaluate the likelihood of each probe object's position $a, b$ under current beliefs: $p(a | x_1, \ldots x_n), p(b| x_1, \ldots x_n).$ The probability of choosing a probe is proportional to the difference in log likelihood:
$$
p(\text{choose } a | x_1, \ldots x_n) \propto \exp (\alpha \cdot(\log p(a | x_1, \ldots x_n) - \log p(b | x_1, \ldots x_n)).
$$

\begin{figure}[t]
    \centering
    \includegraphics[width=0.48\linewidth]{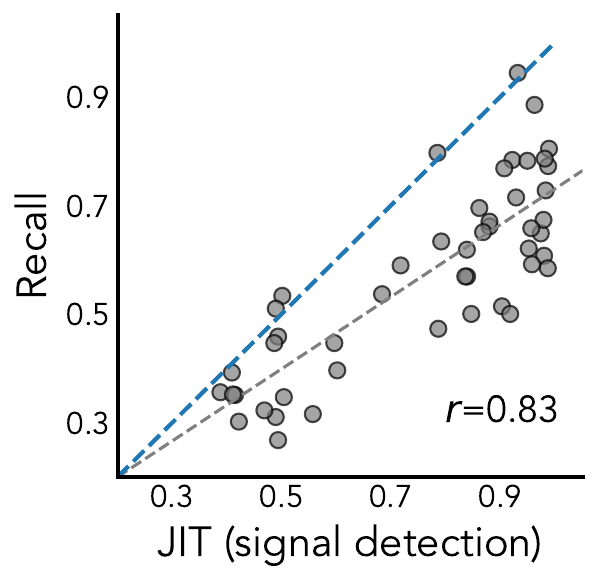}
    \includegraphics[width=0.48\linewidth]{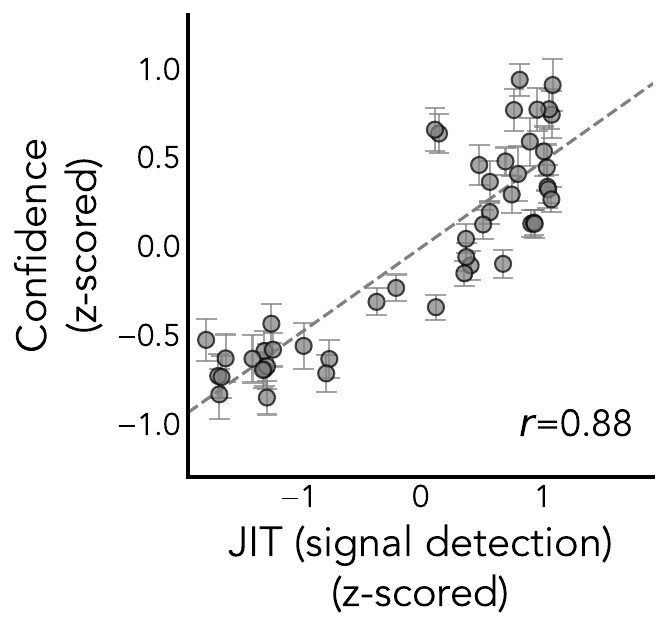}
    \caption{Scatterplots showing predictions from the physics VGC model against Recall (left) and Confidence (right) in physics Experiment 2A.}
    \label{fig:supp-jit-signaldetect}
\end{figure}

Supplemental Figure~\ref{fig:supp-jit-signaldetect} shows predictions from the signal detection variant of the JIT model on recall from Physics Experiment 2A, after fitting the likelihood variance $\kappa$ and the choice temperature $\alpha$. Despite having one extra free parameter on top of the original JIT model, this signal detection variant results in a worse fit overall (recall: $r=0.83$, rmse: $0.22$, confidence: $r=0.88$).

\subsection{Additional Physics Baseline}

We tested an additional baseline that predicted performance on memory probes not as reflecting effects of object representations in working memory, but rather as a counterfactual judgment. More specifically, our \textit{reconstructive baseline} hypothesizes that people remember the predictions they made during the task, formalized as a set of \textit{reference predictions} $q_\text{ref}$, and resimulate during the memory probe to identify the object that produced simulations more closely resembling the predictions they remember.
For each probe object $a, b,$ the model generates candidate solutions $q_a^i, q_b^i$ assuming object $a$ was the real object, and vice versa.
The probability of a choice is directly proportional to the difference between the counterfactual simulations and the reference simulation, measured using the one dimensional wasserstein distance. We normalize wasserstein distances as compared to a completely random prediction $q_\text{uniform}$:
$$
p(a) \propto \exp\left(-\alpha^{-1
} \cdot \frac{W_1(q_a, q_\text{ref})}{W_1(q_\text{uniform}, q_\text{ref})}\right)
$$
We fit the softmax temperature $\alpha$ using grid-search.
This baseline performs worse than the \shortmodel~model, with $(r=0.58, \text{RMSE}=0.27)$ for recall, and $(r=0.63)$ for confidence; see Supplemental Figure \ref{fig:supp-metacognitive-results} for more details.

\begin{figure}
    \centering
    \includegraphics[width=0.49\linewidth]{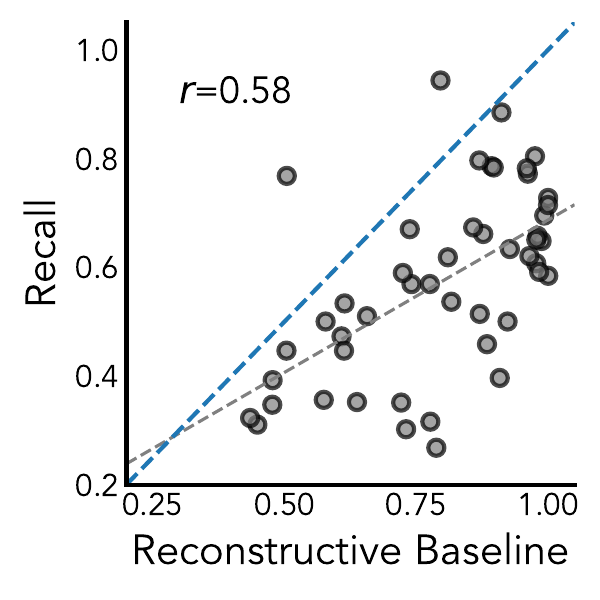}
    \includegraphics[width=0.49\linewidth]{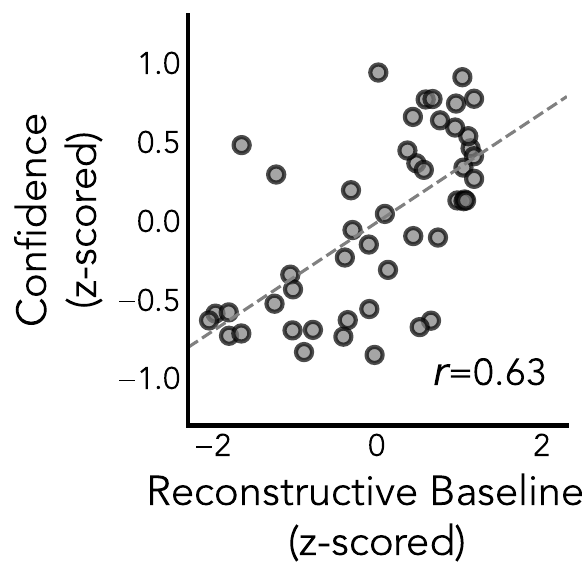}
    \caption{Scatterplots showing the reconstructive baseline's predictions against Recall (left) and Confidence (right) in physics Experiment 2A.}
    \label{fig:supp-metacognitive-results}
\end{figure}

\subsection{Additional control analyses}

Finally, we performed several additional analyses to see how much our results were driven by low-level features of the stimuli we used.

\begin{figure}
    \centering
    \includegraphics[width=0.7\linewidth]{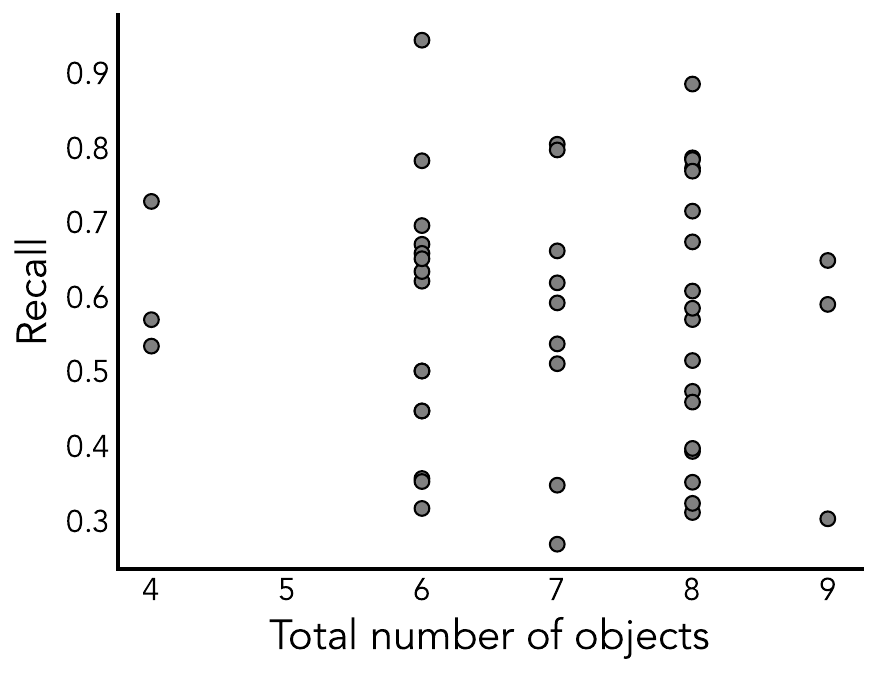}
    \caption{Scatterplots showing correlation between the total world size (measured as the number of objects in the world) on object recall, in physics Experiment 2A.}
    \label{fig:supp-worldsize-control}
\end{figure}

\begin{figure}
    \centering
    \includegraphics[width=0.7\linewidth]{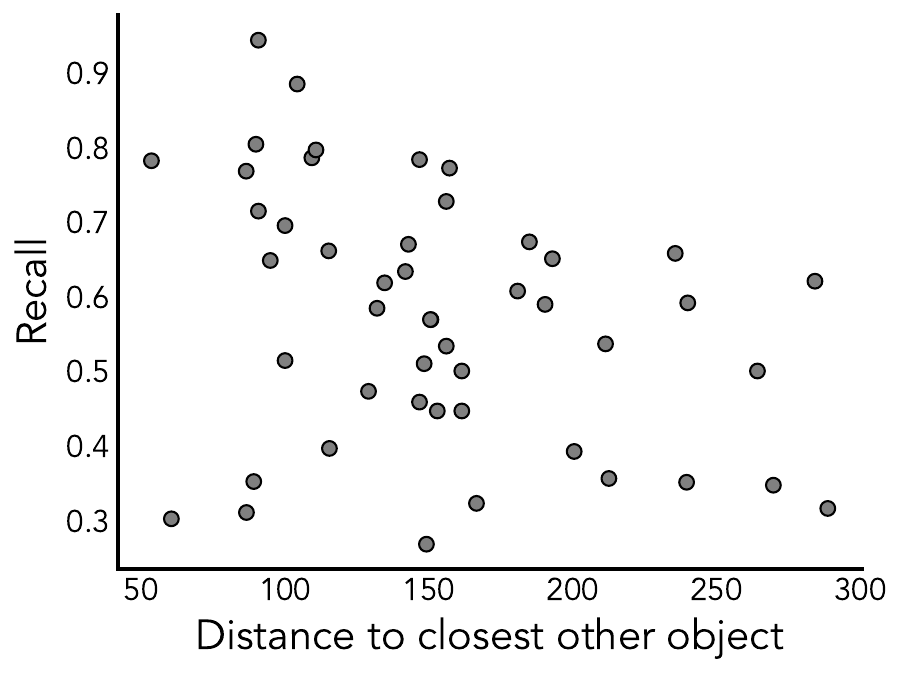}
    \caption{Scatterplots showing correlation between the distance to nearest neighbor on object recall, in physics Experiment 2A.}
    \label{fig:supp-distance-control}
\end{figure}

\begin{figure}
    \centering
    \includegraphics[width=0.7\linewidth]{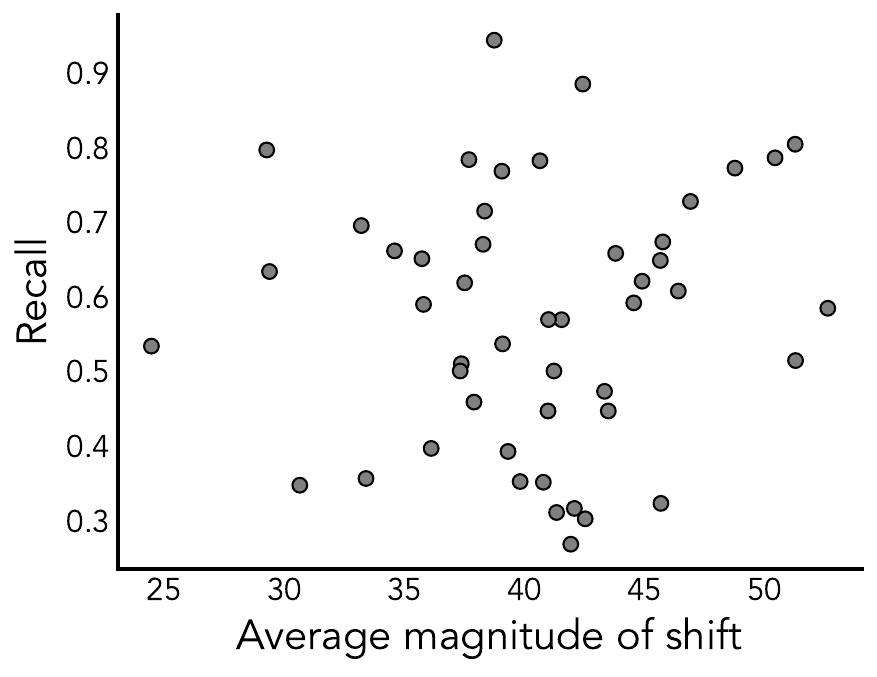}
    \caption{Scatterplots showing correlation between the average distance between target and lure stimuli on object recall, in physics Experiment 2A.}
    \label{fig:supp-shift-control}
\end{figure}

\begin{figure}
    \centering
    \includegraphics[width=0.7\linewidth]{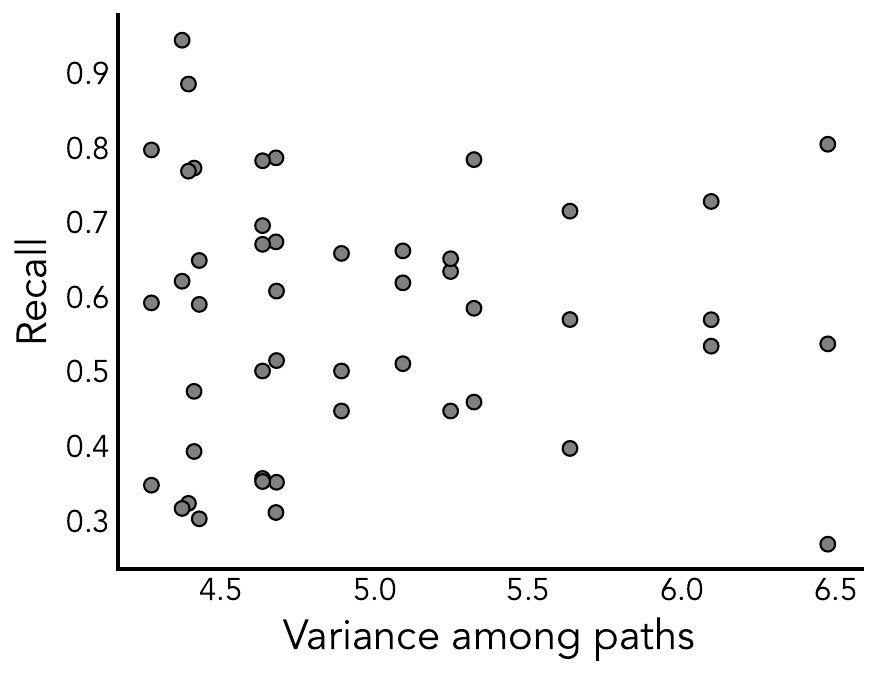}
    \caption{Scatterplots showing correlation between the variance of ball trajectories on object recall, in physics Experiment 2A.}
    \label{fig:supp-midpath-control}
\end{figure}

We analyzed recall performance as a function of: (1) the total number of objects in the world (Supplemental Fig.~\ref{fig:supp-worldsize-control}), (2) proximity of the target object to nearby objects (Supplemental Fig.~\ref{fig:supp-distance-control}), (3) how much the lure stimuli were shifted (Supplemental Fig.~\ref{fig:supp-shift-control}), and (4) the amount of variance in simulated trajectories, measured as the standard deviation of x positions at the midpoint of trajectories (Supplemental Fig.~\ref{fig:supp-midpath-control}.
The size of the world, the magnitude of the shift, and midpath variance were completely uncorrelated to recall ($r=-0.06,0.09,-0.01$) respectively. The proximity to the nearest neighbor showed more explanatory power $(r=-0.33)$, but semi-partial correlations showed it only explained marginally more variance after accounting for the JIT predictions ($r_{x,y}=-0.12$).

\subsection{Reanalysis of planning experiments}
\label{sec:supp-reanalysis-planning}
We fit the parameters of the \shortmodel~model to each of the 9 experiments from \parencite{ho2022people}, including the three experiments shown in the main text.
The full set of results for all 9 experiments are shown in Extended Data Figure \ref{fig:extended-navigation-scatterplots}.
In all experiments, the \shortmodel~model was better correlated with human performance than VGC.

\begin{figure}
    \centering
    \includegraphics[width=0.99\linewidth]{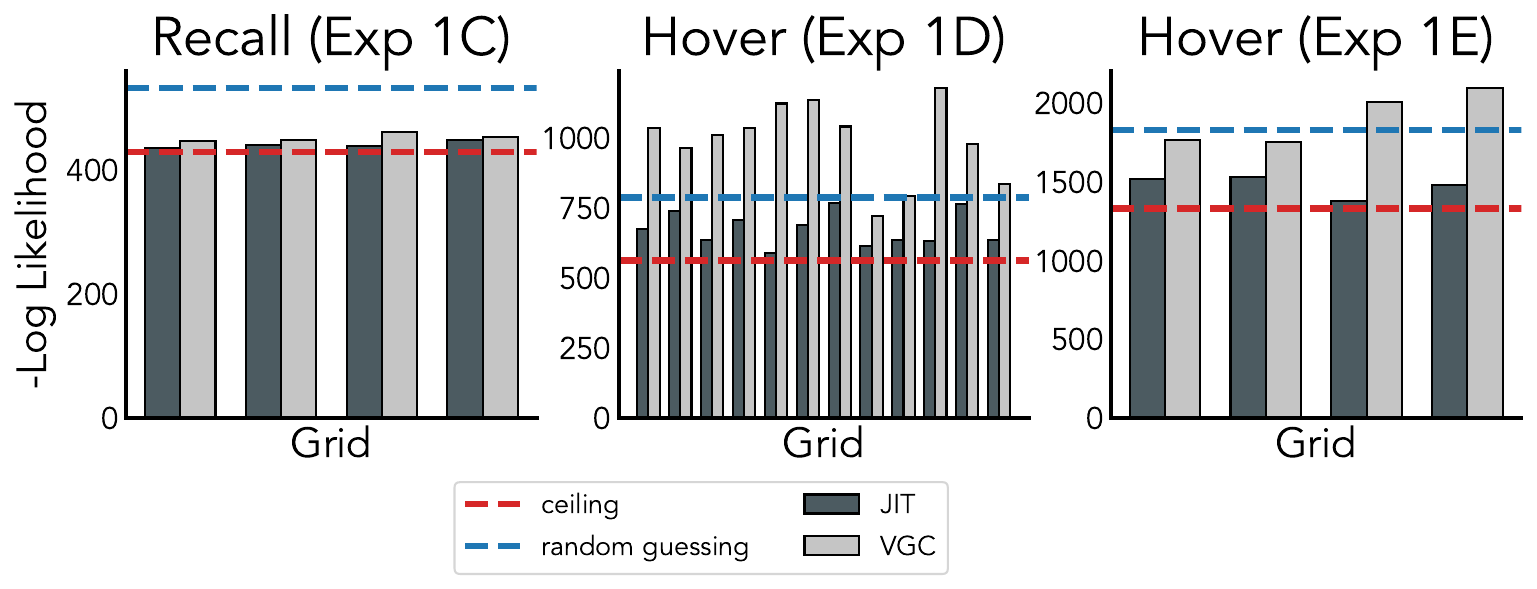}
    \caption{Bar plots showing per-grid likelihoods for the three experiments in the main text. The \shortmodel~model consistently predicts human behavior than VGC across all trials in all experiments.}
    \label{fig:supp-parameter-fits}
\end{figure}

Fitted parameter values are shown in table \ref{tab:supp-parameter-fits}. Note that for the hover experiments, $\gamma$ was fixed to be 0.
Likelihood plots for the three main text experiments are shown in \ref{fig:supp-parameter-fits}, demonstrating that \shortmodel~outperformed VGC across all grids in all three experiments.

\begin{table}[ht]
    \centering
    \begin{tabular}{llll}
        \toprule
        experiment & $\alpha_d$ & $\alpha_h$ & $\gamma$ \\
        \midrule
        initial awareness & 0.00 & 0.05 & 1.34 \\
        upfront awareness & 0.21 & -0.47 & 1.42 \\
        critical memory & 0.00 & -0.21 & 1.11 \\
        critical confidence & 0.21 & -0.21 & 1.42 \\
        critical awareness & 4.00 & 3.74 & 0.79 \\
        initial hover & 0.21 & -0.47 & 0.00 \\
        initial loghover duration & 0.21 & -1.00 & 0.55 \\
        critical hover & 3.37 & 1.37 & 0.00 \\
        critical loghover duration & 0.00 & -0.47 & 0.24 \\
        \bottomrule
    \end{tabular}
    \caption{Best fitting model parameters for each experiment in \parencite{ho2022people}}
    \label{tab:supp-parameter-fits}
\end{table}

\end{document}